\documentclass[lettersize,journal]{IEEEtran}
\usepackage{amsmath,amsfonts}
\usepackage[ruled]{algorithm2e}
\usepackage{array}
\usepackage{textcomp}
\usepackage{stfloats}
\usepackage{url}
\usepackage{verbatim}
\usepackage{graphicx}
\usepackage{subfigure}
\usepackage{bm}
\usepackage{cite}
\usepackage{algpseudocode}
\usepackage{booktabs}
\newcommand\mymodel{PGA}

\usepackage{bbm}
\usepackage{multirow}
\usepackage{enumitem}
\usepackage{hyperref}

\begin{document}

\title{Simple and Efficient Partial Graph Adversarial Attack: A New Perspective}

\author{Guanghui~Zhu,
	    Mengyu~Chen,
        Chunfeng~Yuan,
        and~Yihua~Huang
        % <-this % stops a space
\thanks{Guanghui~Zhu, Mengyu~Chen, Chunfeng~Yuan, and Yihua~Huang are with State Key Laboratory for Novel Software Technology, Nanjing University, Nanjing, China. E-mail: mengyuchen@smail.nju.edu.cn, \{zgh, cfyuan, yhuang\}@nju.edu.cn. Guanghui Zhu is the corresponding author.}
\thanks{Manuscript received April 19, 2021; revised August 16, 2021.}}

\markboth{Journal of \LaTeX\ Class Files,~Vol.~14, No.~8, August~2021}%
{Shell \MakeLowercase{\textit{et al.}}: A Sample Article Using IEEEtran.cls for IEEE Journals}


\maketitle

\begin{abstract}
As the study of graph neural networks becomes more intensive and comprehensive, their robustness and security have received great research interest.
The existing global attack methods treat all nodes in the graph as their attack targets. 
Although existing methods have achieved excellent results, 
  there is still considerable space for improvement. 
The key problem is that the current approaches rigidly follow the definition of global attacks.
They ignore an important issue, i.e., different nodes have different robustness and are not equally resilient to attacks. 
From a global attacker's view, 
  we should arrange the attack budget wisely, 
    rather than wasting them on highly robust nodes.
% To this end, we study the novel global attack method, 
%   which just selects partial nodes as attack targets. 
To this end, we propose a totally new method named partial graph attack (\mymodel{}), which selects the vulnerable nodes as attack targets.
 First, to select the vulnerable items, we propose a hierarchical target selection policy, which allows attackers to only focus on easy-to-attack nodes. 
  Then, we propose a cost-effective anchor-picking policy to pick the most promising anchors for adding or removing edges, and a more aggressive iterative greedy-based attack method to perform more efficient attacks.
 Extensive experimental results demonstrate that \mymodel{} can achieve significant improvements in both attack effect and attack efficiency compared to other existing graph global attack methods.
\end{abstract}


\section{Introduction}
\IEEEPARstart{I}{n} recent years, there has been a surge of interest in studying the robustness of graph neural networks (GNNs), driven by their increasing use in various applications. Many research works on graph adversarial attacks have emerged \cite{NIPA}, \cite{Mettack}, \cite{rls2v}, \cite{SGA}, \cite{GraphFGA}, demonstrating that the addition of just a few adversarial edges, removal of original edges, or inclusion of a few spurious nodes can significantly undermine the performance of GNN models. This vulnerability poses a significant challenge to the widespread adoption of GNNs, despite their impressive performance across various tasks.

One of the main differences between graph adversarial attacks and those in computer vision is the discrete nature of graph structures, which poses unique challenges to developing effective attack methods. However, researchers have made significant progress in this area in recent years. \cite{Nettack} and \cite{rls2v} were among the first to investigate adversarial attacks on graph data, paving the way for subsequent research in this field. As the body of research on graph adversarial attacks has grown, several excellent works that explore various aspects of this problem have emerged. 
For instance, targeted attacks have been studied in depth by \cite{Nettack} and \cite{SGA}. Global attacks, which aim to degrade the performance of an entire graph, have been investigated by \cite{rls2v}, \cite{PGDAttack}, and \cite{RBCDAttack}. Poisoning attacks, which involve manipulating the training data to introduce vulnerabilities into the model, have been studied by \cite{Mettack}. Additionally, \cite{PGDAttack}, \cite{RBCDAttack}, and \cite{GraphFGA} perform evasion attacks, which aim to modify the prediction of GNN models in the test phase. These works have advanced our understanding of graph adversarial attacks and provided important insights into how to develop more robust GNN models.
%

% This paper focuses on global evasion attacks, which aim to significantly decrease the performance of trained GNN models during testing. \cite{PGDAttack} was the first to propose generating adversarial graphs using an optimization approach that maximally disrupts the overall performance, specifically with PGD. Meanwhile, \cite{RBCDAttack} developed PRBCD and GreedyRBCD, which are more efficient and scalable attack methods that leverage random block gradient descent \cite{RBCD} and a carefully designed attack loss. Although existing attack methods have been successful, there is still room for improvement in the field of graph adversarial attacks.
%
This paper focuses on global evasion attacks, which aim to significantly decrease the performance of trained GNN models during testing. The pioneering work~\cite{PGDAttack} introduced the concept of generating adversarial graphs using an optimization approach that maximally disrupts the overall performance, specifically employing the Projected Gradient Descent (PGD) method. Subsequently, \cite{RBCDAttack} proposed PRBCD and GreedyRBCD, which offer more efficient and scalable attacks by leveraging techniques of random block gradient descent \cite{RBCD} and carefully designed attack losses. Despite the successes achieved by existing attack methods, the field of graph adversarial attacks still holds untapped potential for further advancements and improvements.
%

% One key issue with current approaches is that they attempt to modify the classification results for every node in the graph, without considering the variability between nodes. However, we argue that attackers should take into account the differences in the robustness of different nodes under the aggregation paradigm. For example, it is widely agreed that nodes with a large degree are more resistant to interference than nodes with a small degree \cite{HAO}, \cite{Nettack}, \cite{IGJSMA}. By focusing on vulnerable nodes, an attacker with limited budgets $\Delta$ (e.g., the number of changed edges) could improve the success rate and efficiency of the attacks. Unfortunately, existing works have neglected this point.
%

One key issue with current approaches is their tendency to indiscriminately modify the classification results for every node in the graph, overlooking the variability between nodes. However, we argue that attackers should consider the differences in the robustness of individual nodes under the aggregation paradigm. It is widely acknowledged that nodes with a large degree exhibit higher resistance to interference compared to nodes with a small degree \cite{HAO, Nettack, IGJSMA}. By focusing on vulnerable nodes, an attacker with limited resources, such as a restricted budget $\Delta$ (e.g., the number of changed edges), can potentially enhance the success rate and efficiency of the attacks. Unfortunately, existing works have largely overlooked this crucial aspect, failing to incorporate node-level variability into their attack strategies.
%

% Moreover, we borrow the targeted attack method SGA \cite{SGA} and attack each node one by one, recording the required attack budget and plotting the histogram. As we can see in Figure \ref{introduction: targeted attack test}, most of the nodes only need a small number of fake edges to change their classification results. If the attacker prioritizes the budget over these vulnerable nodes, the global attack can be better achieved. 
%

\begin{figure}[t]
  \setlength{\abovecaptionskip}{0.2cm}
  \centering
  \subfigure[Cora]{
  \hspace{-4mm}
  \begin{minipage}[t]{0.48\linewidth}
  \centering
  \includegraphics[width=1.65in]{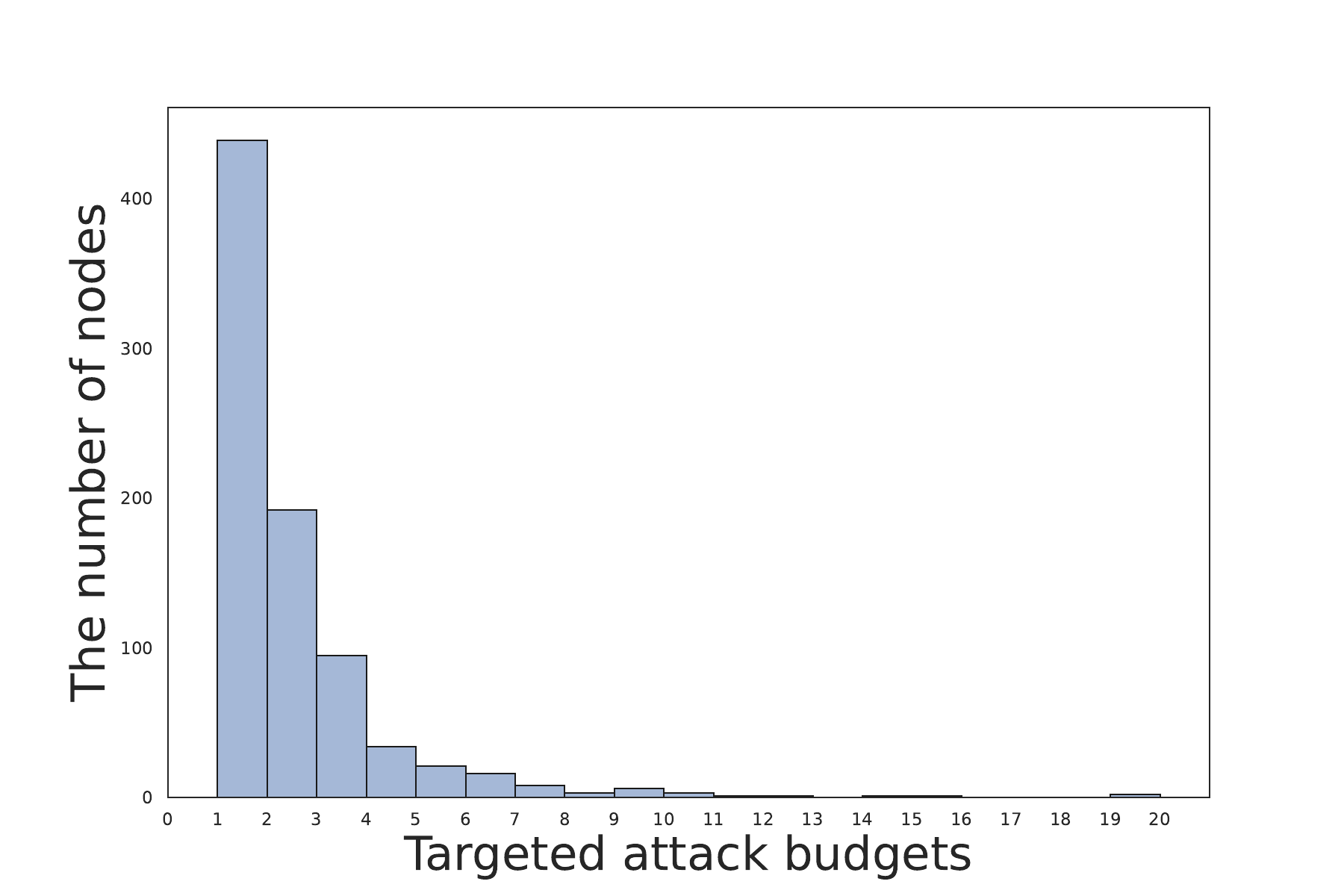}
  %\caption{fig2}
  \end{minipage}
  }%
  \subfigure[Citeseer]{
  \begin{minipage}[t]{0.48\linewidth}
  \centering
  \includegraphics[width=1.65in]{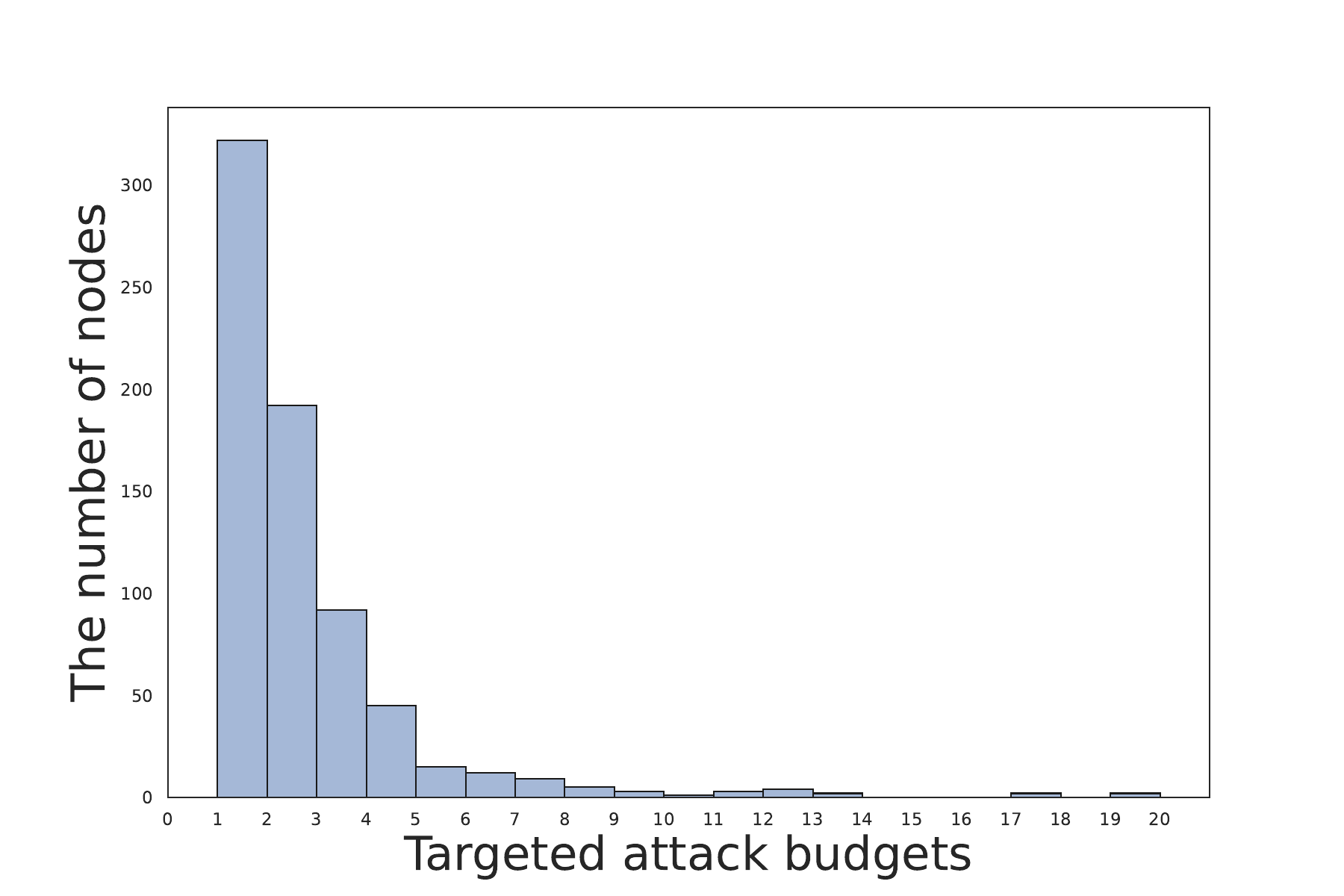}
  %\caption{fig2}
  \end{minipage}
  }%
  \quad
  \centering
  \vspace{-4mm}
  \subfigure[CoraML]{
  \hspace{-4mm}
  \begin{minipage}[t]{0.48\linewidth}
  \centering
  \includegraphics[width=1.65in]{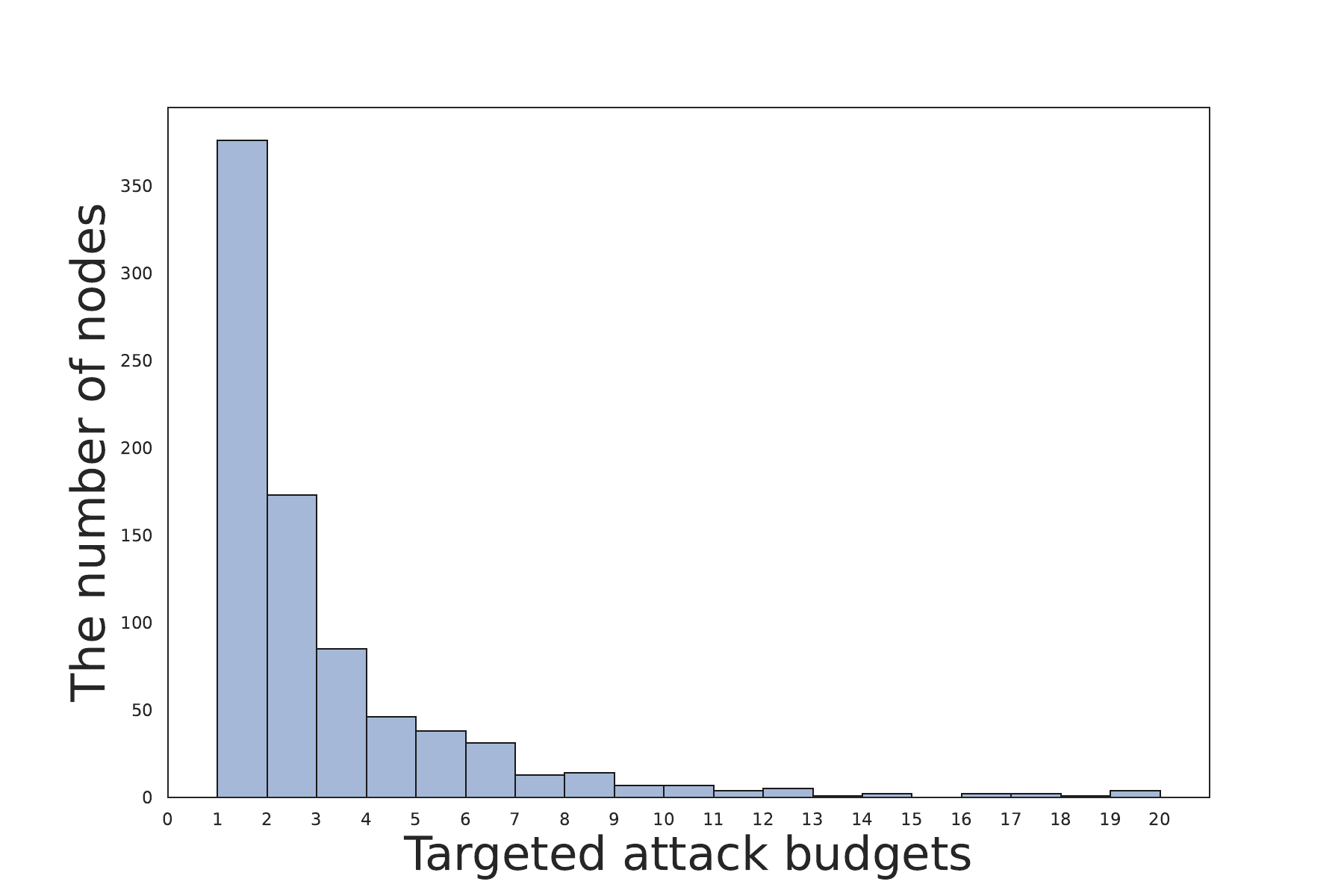}
  %\caption{fig2}
  \end{minipage}
  }%
  \subfigure[Pubmed]{
  \begin{minipage}[t]{0.48\linewidth}
  \centering
  \includegraphics[width=1.65in]{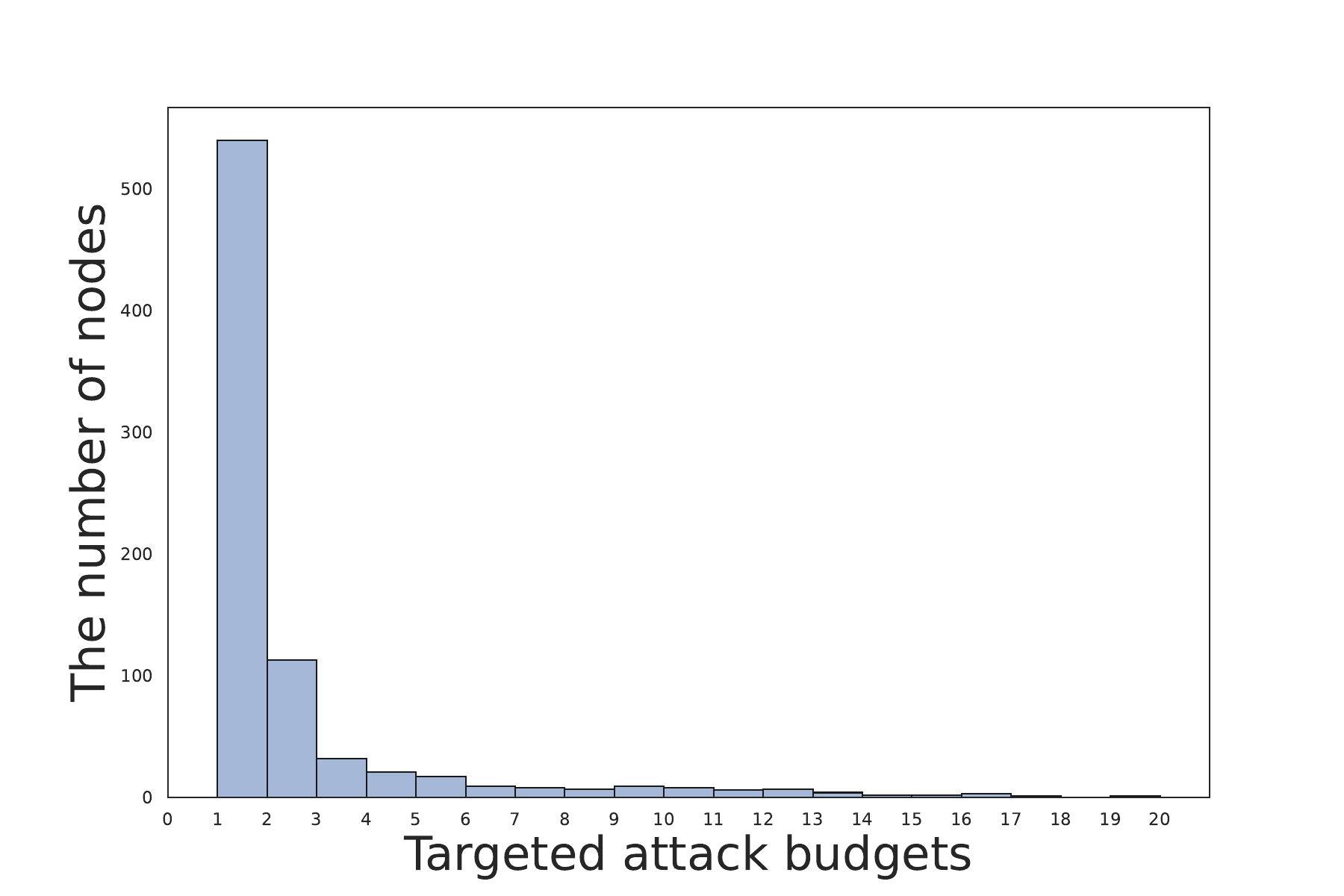}
  %\caption{fig2}
  \end{minipage}
  }%
  \centering
  \vspace{2ex}
  \caption{We individually attack nodes with the targeted attack method SGA and record the corresponding attack budgets required for each node. The x-axis denotes the attack budget. The y-axis denotes the number of nodes that can be successfully attacked given the specific attack budget. }
  \label{introduction: targeted attack test}
  \vspace{-2ex}
\end{figure}

We first adopt the targeted attack method called SGA \cite{SGA} and employ a node-wise approach, attacking each node individually. We record the attack budget required for each node and plot a histogram to visualize the distribution. As depicted in Figure \ref{introduction: targeted attack test}, we observe that the majority of nodes only necessitate a small number of fake edges to alter their classification results. 
The effectiveness of the global attack strategy can be enhanced by prioritizing these vulnerable nodes in terms of the attack budget allocation.

\begin{figure}[t]
    \setlength{\abovecaptionskip}{-0.1cm}
    \centering
    \includegraphics[width=3.25in]{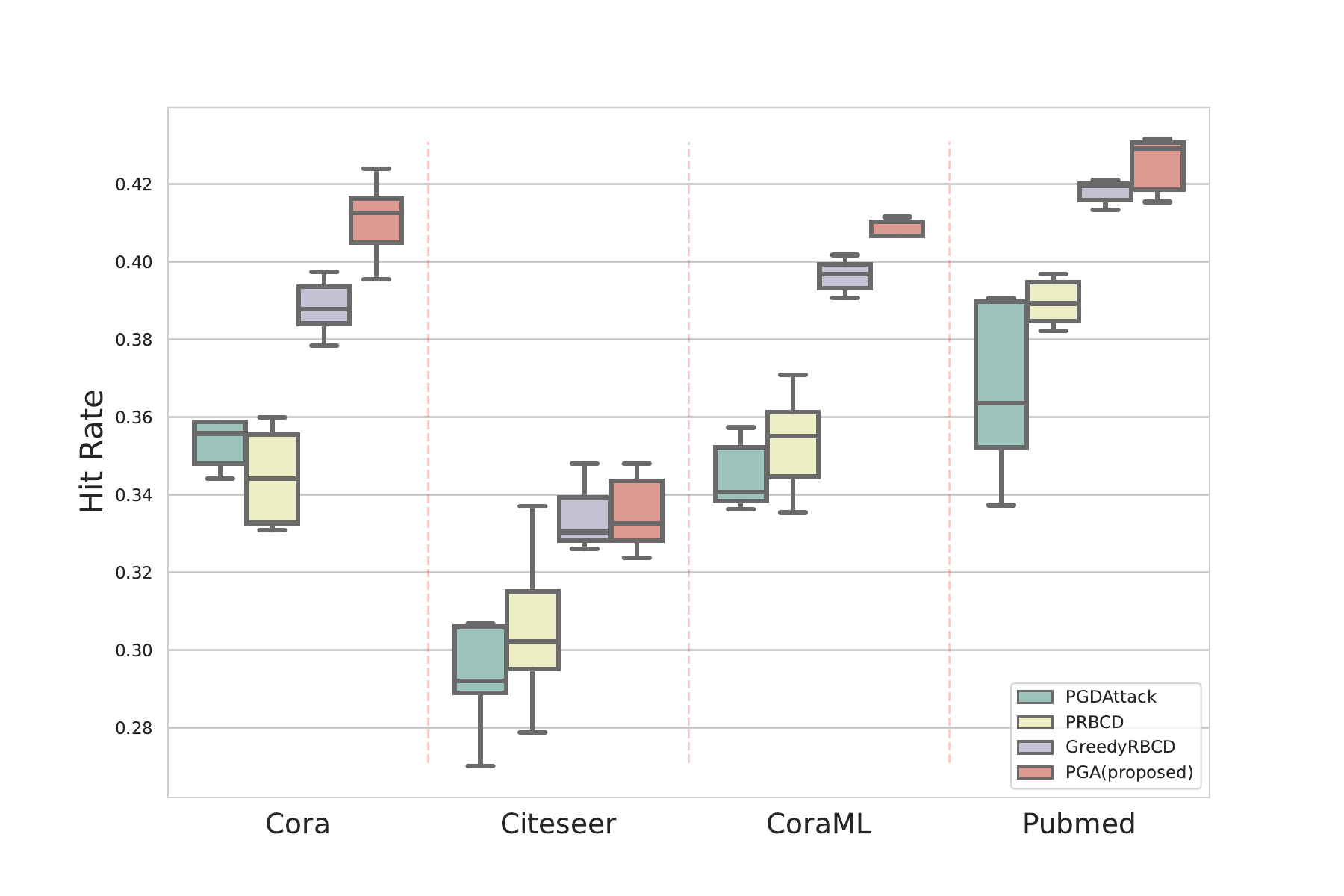}
    \centering
    \caption{A comparison between the proposed \mymodel{} and current global attack methods in terms of the hit rate of vulnerable nodes. We conduct a series of attacks using each method five times.% to ensure reliability in our evaluations. 
    }
    \label{introduction: vulnerable hit rate}
   \vspace{-2ex}
\end{figure}

Drawing from the preceding discussion, we employ the hit rate of vulnerable nodes as a metric to evaluate different attack algorithms. This metric quantifies an attacker's effectiveness in selectively targeting vulnerable nodes. To compute the hit rate of vulnerable nodes, we focus on the malicious edges added by the attacker that directly connect to the most vulnerable nodes. According to \cite{SGA}, direct attacks are shown to possess significantly higher destructive power compared to indirect attacks that focus on influence. These vulnerable nodes are identified as the ones requiring only one fake neighbor for their classification to be modified in a targeted attack. By determining the number of such connections and dividing it by the total attack budget, we obtain the hit rate of vulnerable nodes. As shown in Figure \ref{introduction: vulnerable hit rate}, we observe that existing attack methods have scope for improvement in their capacity to concentrate on vulnerable nodes. 

\begin{figure}[t]
    \setlength{\abovecaptionskip}{-0.1cm}
    \centering
    \includegraphics[width=3.25in]{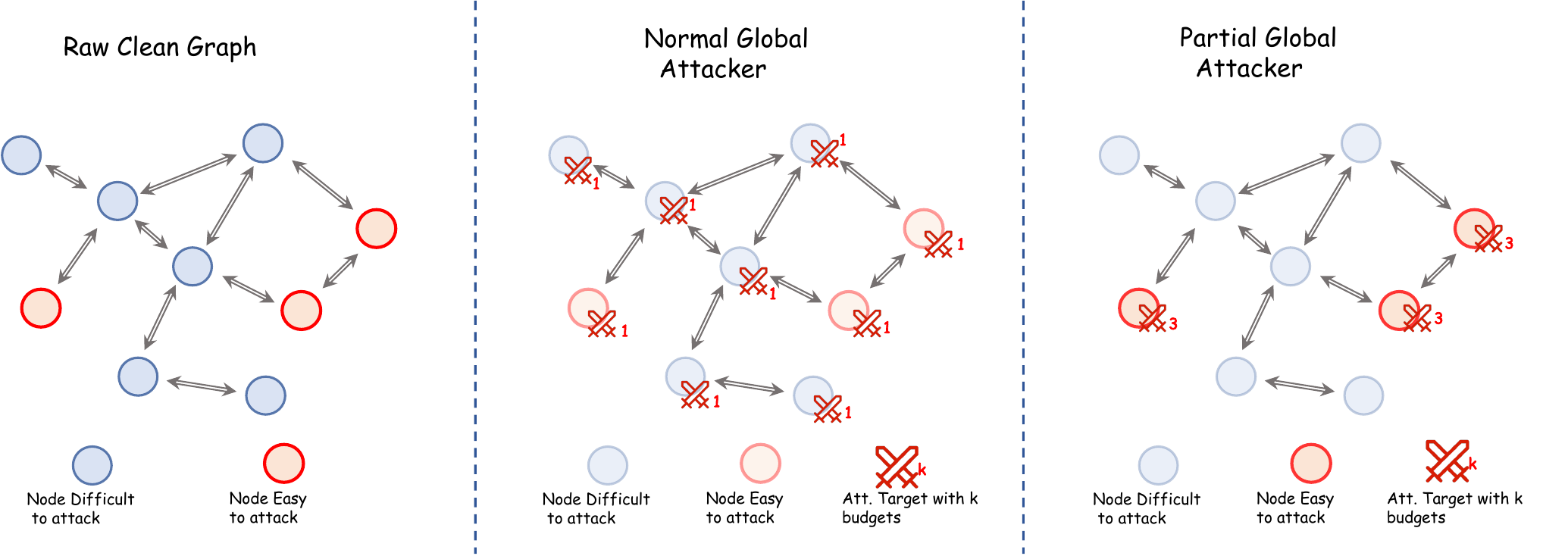}
    \centering
    \caption{Difference between \mymodel{} and existing global attack methods. Assume that the attacker can modify 9 edges, i.e., the attack budget is 9. 
    %The comparison aims to evaluate the performance and efficacy of \mymodel{} in relation to traditional global attack methods.
    }
    \label{introduction: comparison}
    \vspace{-3ex}
\end{figure}

Based on these observations, we assert that it is crucial to fully consider the variations in node robustness when allocating the attack budget. In light of this, we propose a novel global attack method on graph data called Partial Graph Attack, denoted as \mymodel{}\footnote{\mymodel{} is available at https://github.com/PasaLab/PGA}. To illustrate the effectiveness of \mymodel{}, we provide a comparison between \mymodel{} and traditional global attack methods in Figure \ref{introduction: comparison}. The comparison highlights the distinctive characteristics and advantages of our proposed approach.
The normal global allocate the attack budget to each node equally. 
In contrast, the partial global attacker allocates more budgets for vulnerable nodes. 
% Based on these observations, we argue that the differences in node robustness should be fully considered by allocating the attack budget to vulnerable nodes. Thus, we propose a novel global attack method on graph data, called partial graph attack, namely \mymodel{}. A comparison of \mymodel{} and traditional global attack methods is shown in Figure \ref{introduction: comparison}. 

% Specifically, we first propose the hierarchical target selection policy module to find out those fragile nodes as our attack targets, which is completely different from the previous method of attacking all items. Among the target selection policy, the preprocessing filter is introduced to roughly screen out nodes already misclassified by the victim models using the output logits. Next, the degree and margin filters are used to filter out vulnerable nodes. 
%
Specifically, we introduce a \emph{hierarchical target selection policy} module as a fundamental component of our approach, distinguishing it from previous methods that indiscriminately attack all nodes. The module focuses on identifying fragile nodes as our attack targets. Within the target selection policy, we incorporate a preprocessing filter that leverages output logits to identify nodes already misclassified by the victim models, allowing for their exclusion from the attack. Additionally, we utilize degree and margin filters to identify and filter out vulnerable nodes based on their degree centrality and classification margin, respectively. This hierarchical target selection process ensures a more precise and effective selection of nodes for our attacks.
%

% Subsequently, we propose a cost-effective anchor-picking policy to pick anchors for the selected attack targets. Anchors indicate fake neighbors to add or actual neighbors to remove. The anchor-picking policy considers only the most promising items to reduce the number of anchors explored. Also, to further reduce the computational overhead, it employs a surrogate model to compute the gradient once and then keeps only the part of the anchors that are most effective in reducing the attack loss. This approach dramatically improves efficiency without the loss of attack effect. 
%
Subsequently, we introduce a \emph{cost-effective anchor-picking policy} to select anchors for the identified attack targets. Anchors serve as indicators for either adding fake neighbors or removing actual neighbors. Our anchor-picking policy focuses on exploring the most promising items, thereby reducing the number of anchors to be considered. Additionally, to alleviate computational overhead, we utilize a surrogate model to compute the gradient once and retain only the most effective subset of anchors in terms of reducing the attack loss. This approach significantly enhances efficiency without compromising the effectiveness of the attack.
%

% Moreover, we propose an iterative greedy-based attack module to generate the adversarial graph. The number of iterations is set according to the attack budget and in each iteration, the module only uses the calculated edge gradients to decide which edge to add or which edge to remove with a greedy strategy. To further accelerate the generation of the adversarial graph, we introduce a more aggressive edge-flip strategy, which enables the attacker to flip multiple edges after a single gradient calculation.
%
Furthermore, we propose an \emph{iterative greedy-based attack} module for generating the adversarial graph. The number of iterations is determined based on the attack budget, and in each iteration, the module utilizes calculated edge gradients to make decisions on which edges to add or remove using a greedy strategy. To expedite the adversarial graph generation process, we introduce a more aggressive edge-flip strategy, enabling the attacker to flip multiple edges following a single gradient calculation. This approach significantly accelerates the generation of the adversarial graph while maintaining attack effectiveness.
%

% The main contributions of our work are summarized as follows:
% \begin{itemize}
%   \item[$\bullet$] 
%   We propose a novel partial graph attack method, namely \mymodel{}, to improve attack effect and attack efficiency. 
%   To the best of our knowledge, it is the first method to directly attack the vulnerable items 
%     instead of all the nodes in the domain of global graph attack;
%   \item[$\bullet$]
%   To select the vulnerable items, we propose a hierarchical target selection policy, which allows attackers to only focus on easy-to-attack nodes;
%   \item[$\bullet$]
%   We use a cost-effective anchor-picking policy to pick the most promising anchors for adding or removing edges, which greatly reduces the anchor search space, and a more aggressive iterative greedy-based attack module to perform more efficient attacks.
%   \item[$\bullet$]
%   We conduct extensive experiments on five datasets by attacking six widely adopted graph neural networks. The experiment results demonstrate that our novel attack method has achieved significant improvements in both attack effect and attack efficiency compared to other existing graph global attack methods.
% \end{itemize}

The main contributions of our work can be summarized as follows:
\begin{itemize}
\item[$\bullet$] We propose \mymodel{}, a novel partial graph attack method, which enhances both attack effectiveness and efficiency. Notably, it is the first approach to specifically target vulnerable nodes rather than attacking all nodes in the global graph attack domain.
\item[$\bullet$] To identify vulnerable nodes, we introduce a hierarchical target selection policy, which enables attackers to focus on nodes that are easy to attack..
\item[$\bullet$] We introduce a cost-effective anchor-picking policy, which selects the most promising anchors for adding or removing edges, significantly reducing the search space and enhancing attack efficiency. Additionally, we employ a more aggressive iterative greedy-based attack module to further expedite the attack process.
\item[$\bullet$] We conduct extensive experiments on five datasets, targeting six widely adopted graph neural networks. The experimental results demonstrate that our novel attack method outperforms existing graph global attack methods in terms of both attack effectiveness and efficiency.
We also believe that
such a graph adversarial attack method from a new perspective
could make further progress in the future.
\end{itemize}

\section{RELATED WORK}
\subsection{Graph Adversarial Attack} 
In recent years, the robustness of graph neural networks has received increasing attention from researchers. \cite{Nettack} and \cite{rls2v} first start to study adversarial attacks on graph data. \cite{rls2v} focuses on the global evasion attacks for the graph classification and node classification tasks. \cite{Nettack} focuses on targeted attacks on node classification tasks. 
The evasion attack is an attack after the model has been trained, 
as opposed to the poisoning attack \cite{Mettack}. The targeted attacker attacks a single node once, such as \cite{Nettack, rls2v, SGA, IGJSMA}, as opposed to the global attack, which tries to decrease the overall performance of the victim GNN models, such as \cite{Mettack}, \cite{Rewiring}, \cite{PGDAttack}, \cite{RBCDAttack}. Commonly used methods of graph data perturbation 
include adding and removing edges \cite{PGDAttack}, modifying node features \cite{IGJSMA}, and adding fake nodes \cite{TDGIA}.

PGD\cite{PGDAttack} relaxes discrete edge types into continuous types from an optimization perspective, then solves the optimization problem, and finally randomly samples back to discrete edge types. \cite{RBCDAttack} addresses the scalability problem of \cite{PGDAttack} by introducing random block coordinate gradient descent \cite{RBCD}, and proposes PRBCD and its variant GreedyRBCD. 
In addition, two novel attack loss functions are designed to solve the problem of \cite{PGDAttack}, which imposes the attack budgets on the already misclassified nodes.

\subsection{Potential Enhancement Possibility}
Although existing methods achieve excellent performance on global attacks, there is still potential for enhancement. Global attackers try to disrupt the overall performance of the GNN models. Thus, they need to consider all items when calculating the attack loss. 
But they all ignore the fact that different items have different properties. Some items may be more robust, while some may be very fragile. If the limited attack budgets are allocated to the more vulnerable nodes, it is likely that the effectiveness and efficiency of the attacks can be significantly improved.

\subsection{Difference between Counterparts and \mymodel{}}
The key idea of this work is to perform partial graph attacks, which give all the attack budgets to several vulnerable nodes. To the best of our knowledge, our proposed \mymodel{} is the first attack approach that takes into account differences in the robustness of nodes. \mymodel{} directly selects vulnerable nodes as the attack targets. To some extent, the attack loss of \cite{RBCDAttack} is subtly designed to select the attack targets but is still very different from our approach. 

Our work falls in the area of graph global evasion attack, which aims at attacking the models to decrease overall performance in the test phase, by adding or deleting edges. In this work, we focus on the node classification task.

\section{PROBLEM DEFINITION}

We consider the task of (semi-supervised) node classification task. 
Let $\mathcal{G}=(V, A, X)$ be an attributed graph, where $V=\{v_1, v_2, \dots, v_N\}$ denotes the set of nodes, $A \in \{ 0, 1 \}^{N \times N}$ is the adjacency matrix and $X \in \mathbf{R}^{N \times D}$ is the feature matrix. $V_L$ is the labeled node set and $V_U$ is the unlabeled node set with $V_L \cup V_U = V$. The goal of the semi-supervised node classification task is correctly labeling the unlabeled nodes in $V_U$. The model trainer uses nodes in $V_L$ to train the graph node classifier $f$.

\subsection{Attacker's Goal}
In this work, we aim at decreasing the classification accuracy of the trained model for the node classification task in the test phase. The model is fixed, and the attacker cannot access or change the model parameters and the mode structure. Currently, we focus on evasion attacks, while poisoning attacks can also be done without additional changes. In the experimental section, we also conduct several experiments to the effectiveness of \mymodel{} in the context of poisoning attacks.
%

% \subsection{Attacker's Knowledge}
% Our work belongs to the white-box attack. 
% The gray-box attacker only has limited knowledge 
%  about the victim models \cite{jin2021adversarial}. 
% We know nothing about the classification model and
% its trained weights, but we know everything the model trainer knows, including the feature matrix, adjacency matrix, and $V_L$. 
% Although the attacker
% does not know the parameters and structural information of the victim model, 
% a surrogate model can be trained to guide perturbations \cite{zugner2018adversarial}, \cite{https://doi.org/10.48550/arxiv.1902.08412}. 
% In this work, we train a surrogate model to simulate the victim 
%   and get useful information from it to give us gradient information 
%   for adding or removing edges.
\subsection{Attacker's Capability}
Among adversarial attacks, the attacker often has to satisfy the imperceptible properties. Adversarial attacks in the domain of computer vision are on continuous space, e.g. the range of perturbation cannot exceed 8 per pixel.
Besides, the image sample is generally i.i.d. 
Finally, the imperceptible properties of image perturbations can be seen very visually by the eyes. Graph data are very different. It is impossible to have an intuitive understanding of the perturbations visually. Another problem is discretization, which makes the gradient descent method not directly usable. \cite{Nettack} uses budget constraint and node degree distribution to measure the imperceptible properties of graph data perturbations.
Also, due to computational complexity, the budget constraint is widely used \cite{PGDAttack, RBCDAttack, Mettack}, e.g. a limitation that the number of modified edges by attackers cannot exceed 5\% of the total.

\section{THE PROPOSED METHOD}
\subsection{An Overview of the Proposed Method}

\begin{figure*}[h]
    \includegraphics[width=\textwidth]{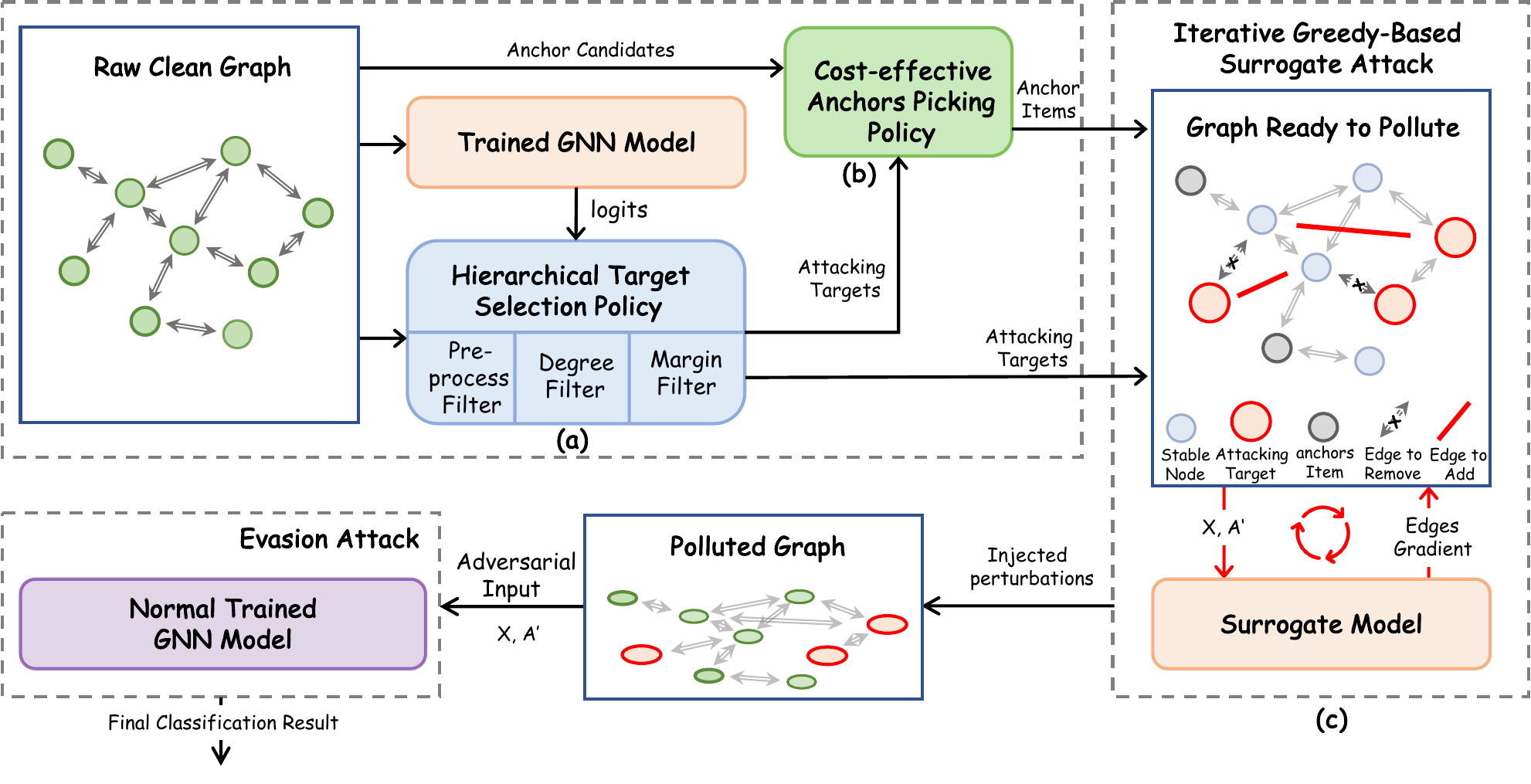}
    % \vspace{-6mm} 
    \vspace{-3ex}
    \caption{
      An overview of the proposed \mymodel{}. 
      \mymodel{} first needs a trained GNN model to generate prediction logits for all nodes. 
      Then, given the logits, 
        the hierarchical target selection policy selects partial nodes as attack targets (a).
      Then, according to attack targets selected by (a), 
        the anchor-picking policy picks 
        the most promising anchors for 
        adding or removing edges between them and attack targets (b). 
      After (a) and (b),  the iterative attack module uses a surrogate model to calculate gradients for each fake edge, 
        and then flips the edge with the largest gradient iteratively, until the attack budgets are exhausted (c). 
      Finally, a polluted graph is generated as the adversarial input for an evasion attack.
      }
    % \vspace{-2mm}
    \label{framework}
  \end{figure*}
The goal of this work is to conduct global attacks on a trained graph model. To achieve the goal more efficiently and increase the overall success rate of attacks, we propose a partial graph attack method \mymodel{}, which concentrates budgets on the nodes easy to be disturbed. The overall framework of the proposed method \mymodel{} is shown in Figure \ref{framework}, which consists of three main components: hierarchical target selection policy (a), cost-effective anchor-picking policy (b), and iterative greedy-based surrogate attack (c). To select attack targets to perform partial attacks, we use a hierarchical target selection policy, which has three steps: 
  (1) localizing misclassified items (Preprocessing Filter), 
  (2) filtering out items with a large degree (Degree Filter), and (3) dropping out items with a large margin (Margin Filter). 
Then, a cost-effective anchor-picking policy is introduced to pick the most promising anchors for changing the predicted labels of the selected attack targets. Fake edges may be added between anchors and the attack targets, or existing edges may be removed between them. Finally, with the help of target selection policy and anchor-picking policy, the last component iteratively adds fake edges or removes existing edges between targets and anchors for generating an adversarial graph.

\subsection{Hierarchical Target Selection Policy}
\label{ss: hierarchical}
According to the setting of the proposed partial graph attack, we need to select the set of victim nodes to be attacked first. 
To this end, we design a hierarchical target selection policy, which aims to select nodes that may be easier to attack. 
In this section, we first analyze which characteristics of the node itself are most likely to be related to its robustness, and then introduce our hierarchical target selection policy.

\subsubsection{Statistics-Robustness Relationship Analysis}
We aim to leverage certain model-independent node statistics to predict their robustness before conducting attacks. 
Node degree is widely regarded as a viable indicator, as mentioned in \cite{Nettack}, \cite{Mettack}, and \cite{TDGIA}. 
It is believed that nodes with low degrees are more susceptible to the influence of fake neighbors. 
Additionally, the classification margin of a node is directly associated with the correctness of its classification result. 
Taking binary classification as an example, if a sample's classification margin is small, it suggests that it is situated near the classification boundary.
Consequently, when the sample is perturbed by noise, it is more prone to crossing the boundary and altering its classification result. 
Moreover, we consider other node characteristics, such as PageRank score \cite{pagerank}, clustering coefficient \cite{cluster_coeff}, and eigenvector centrality \cite{eigen_centrality}. 
We exclude certain statistics that require excessive computation time, such as betweenness centrality.

Firstly, we employ PGDAttack \cite{PGDAttack} to execute an attack and identify two types of nodes: (1) nodes that were originally classified correctly by GCN but incorrectly classified after the attack, and (2) nodes that GCN could classify correctly both before and after the attack. We denote these two types of nodes as two classes. 
Consequently, we formulate a binary classification task, aiming to predict whether a sample will be successfully attacked based on its attributes.
We consider attributes such as degree, PageRank score, eigenvector centrality, clustering coefficient, and classification margin.  
Our objective is to investigate the significance of different attributes. 
Note that we only employ PGDAttack for 
the purpose of analysis, and the actual execution of \mymodel{} attack does not rely on any other attack methods.

\begin{figure}[t]
  % \vspace{8mm}
  \subfigure{
    \setlength{\abovecaptionskip}{-0.1cm}
    \centering
    \includegraphics[width=3.25in]{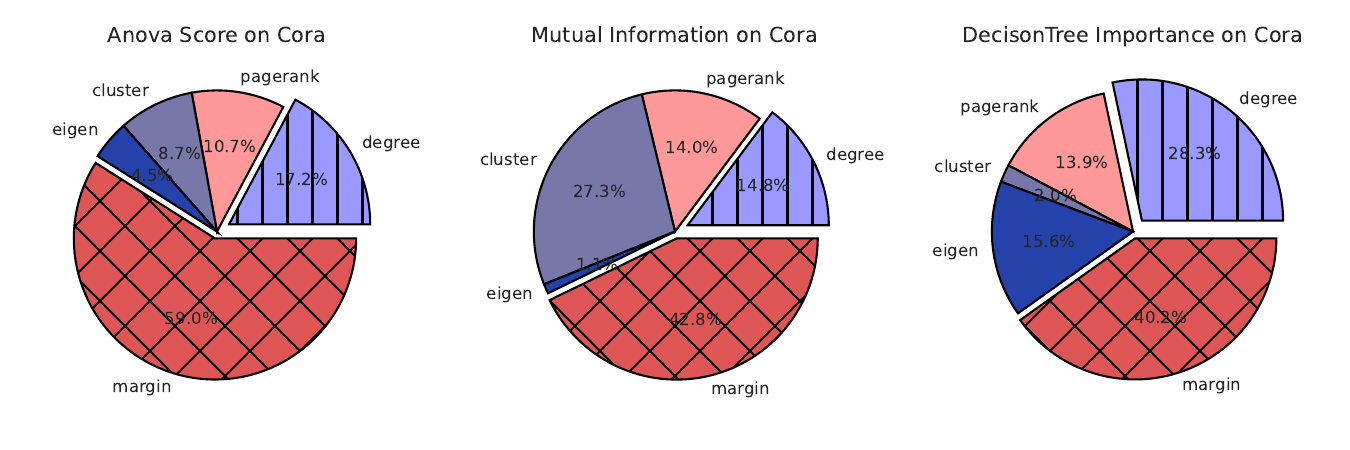}
  }
  \subfigure{
    \setlength{\abovecaptionskip}{-0.1cm}
    \centering
    \includegraphics[width=3.25in]{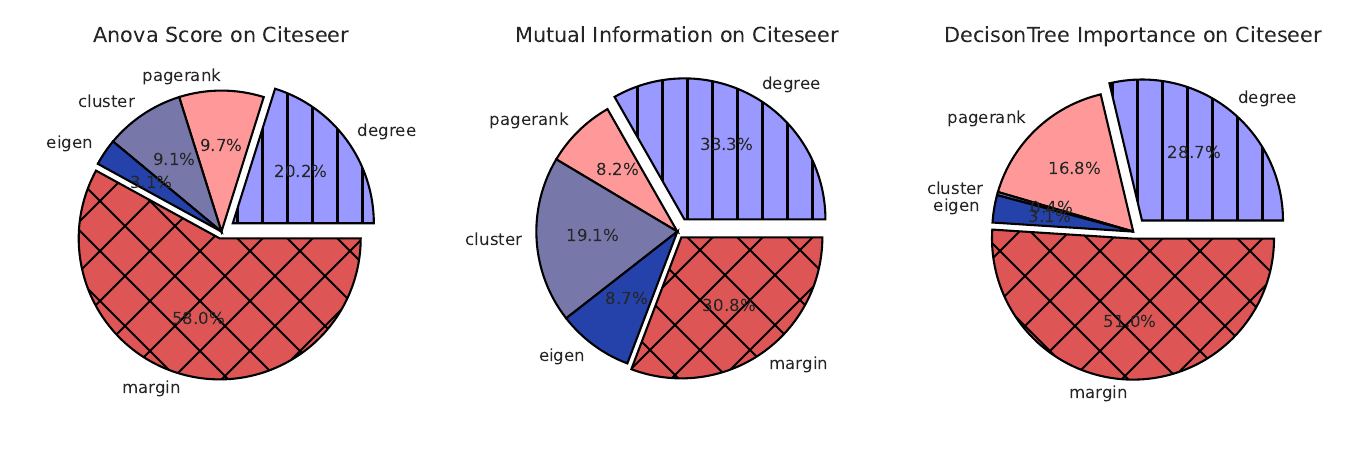}
  }
  \subfigure{
    \setlength{\abovecaptionskip}{-0.1cm}
    \centering
    \hspace{-3mm}
    \includegraphics[width=3.5in]{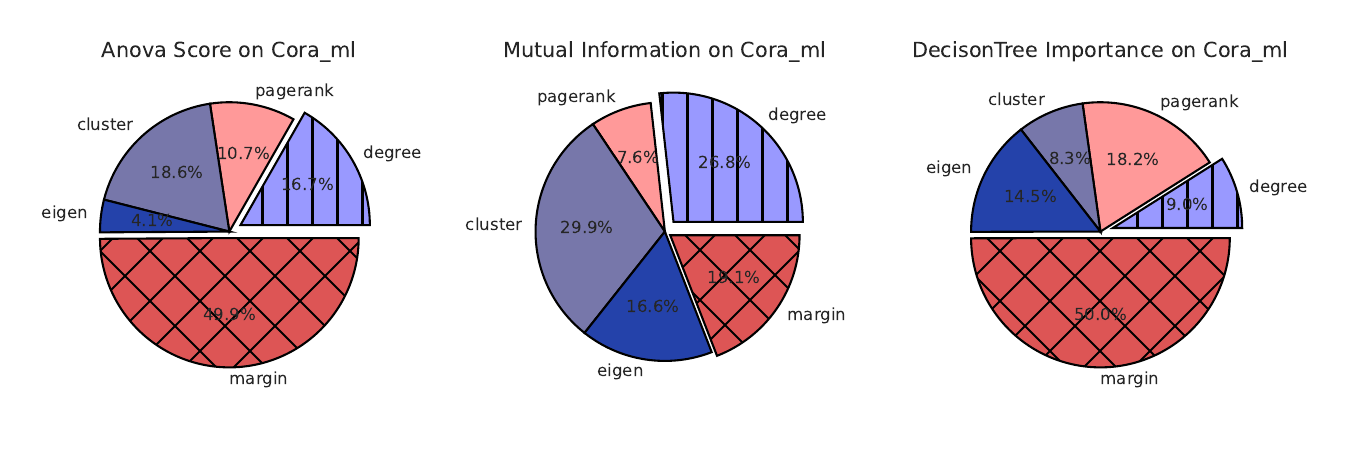}
  }
  \subfigure{
    \setlength{\abovecaptionskip}{-0.1cm}
    \centering
    \hspace{-4mm}
    \includegraphics[width=3.5in]{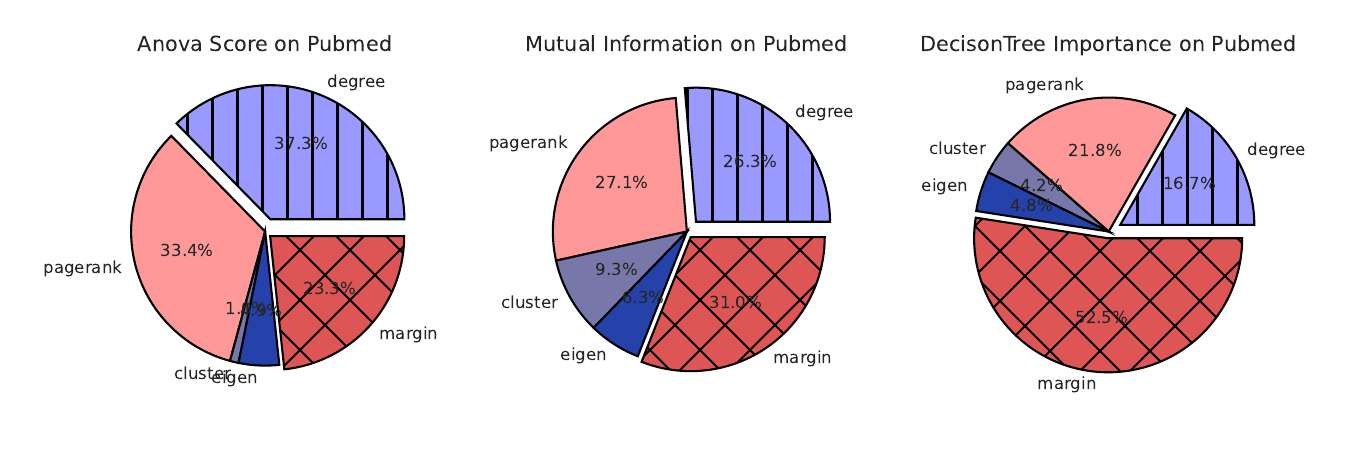}
  }
  %\vspace{1ex}
  %\captionsetup{singlelinecheck=off,justification=centering}
  \caption{Results on statistics-robustness relationship analysis. Degree and classification margin are more correlated to the node robustness.}
  \label{fig: feature selection}
  \vspace{-2ex}
\end{figure}

Next, we employ three different methods for feature importance analysis: decision tree analysis \cite{decision_tree}, ANOVA analysis \cite{ANOVA}, and mutual information analysis \cite{mutual_infos}. 
To perform these analyses, we utilize the well-established third-party library scikit-learn \cite{Scikit_learn}. 
Once we obtain the importance score for each feature, we apply normalization processing. 
The analysis results are presented in Figure \ref{fig: feature selection}.
It is evident that in most cases, the classification margin and degree exhibit the highest importance as characteristics.
In summary, when considering the statistics of individual nodes, both degree and classification margin demonstrates the strongest correlation with their robustness.
Thus, we %have ultimately decided to 
utilize degree and classification margin as indicators for identifying vulnerable nodes. 

\begin{figure*}[h]
  \subfigure[F1 score]{
    \centering
    \includegraphics[width=\textwidth]{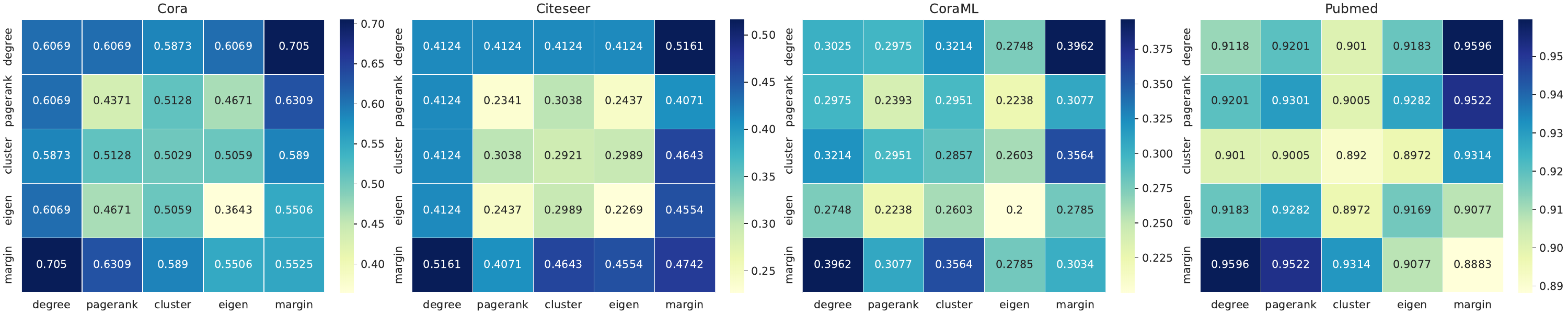}
  }
  \vspace{-4mm}
  \subfigure[AUC]{
    \centering
    \includegraphics[width=\textwidth]{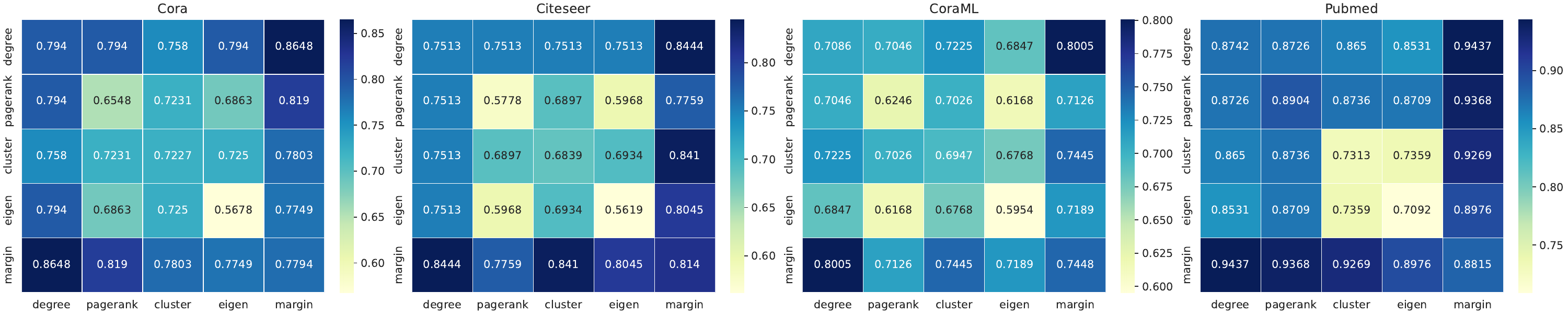}
  }
  \vspace{1ex}
 % \captionsetup{singlelinecheck=off,justification=centering}
  \caption{We utilize the pairs of all available statistics as input for the model to predict the robustness of nodes. F1 score and AUC for the binary classification task are depicted. 
  The combination of degree and classification margin yields optimal results in most cases.
  }
  \label{fig: combination analysis}
  \vspace{-2ex}
\end{figure*}

Since we opt for the combination of degree and classification margin, we proceed to explore additional potential combinations.
As previously mentioned, our binary classification task involves attributes such as degree, PageRank score, clustering coefficient, eigenvector centrality, and classification margin. To explore different combinations, we examine pairs of these statistics. 
We first divide the generated dataset using stratified sampling, allocating 66\% as the training set and the remaining as the validation set. 
For classification purposes, we train a support vector machine classifier (other classifiers can also be used). Finally, we evaluate the classification results by measuring the F1 score and AUC. 
From Figure \ref{fig: combination analysis}, we observe that, in most cases, the chosen combination of degree and classification margin yields optimal outcomes.

\subsubsection{Preprocessing Filter}
In the case where the attack budgets are constrained and fixed, to maximize the damage to the overall performance of the graph model, it is reasonable to focus on the nodes that can be correctly classified by the graph model \cite{revisiting23}. Therefore, it is desirable to first find out items already classified correctly. Mathematically, the preprocessing filter can be defined as follows: 
\begin{equation}
  \mathcal{P}_{flt}(V) = \{ v | \hat{y}_{v} = y_{v} ,~v \in V_U \}
\end{equation}
where $\hat{y}_{v}$ is the model prediction of $v$. $y_{v}$ denotes the true label of $v$. Since the attacker has no access to the true labels of $V_U$, we use the output classification margin to evaluate the possibility of being correctly classified. 
% 与鲁棒性分析不同，我们不认为结点统计量，如degree, eigenvector centrality等，与GNN模型是否可以在该结点上正确分类存在必然联系。所以我们最后仅仅使用了margin来作为筛选的依据。
%
The classification margin (CM) of node $v$ is denoted as follows:
\begin{equation}
  \text{CM}(v) = \mathbf{\tilde{z}}_{v}^{0}- \mathbf{\tilde{z}}_{v}^{1}
\end{equation} 
where $\mathbf{\tilde{z}}_{v}$ denotes the sorted output logit of $v$ in descending order. $\mathbf{\tilde{z}}_{v}^{i}$ denotes the $i$-th element in  $\mathbf{\tilde{z}}_{v}$.  Thus, $\mathbf{\tilde{z}}_{v}^{0}$ and  $\mathbf{\tilde{z}}_{v}^{1}$ represent the first two largest model prediction component of the logit. 
The classification margin can be a measure of classification confidence. A smaller margin means that the GNN model is more unsure of the prediction, and the true label is more likely to be different from the model prediction. 
The preprocessing filter can be finally defined as follows: 
\begin{equation}
  \mathcal{P}_{flt}(V) = \{ v | \text{CM}(v) > \text{threshold}_p ,~v \in V_U \}
\end{equation}

\subsubsection{Degree Filter}

\begin{figure}[h]
    \setlength{\abovecaptionskip}{-0.1cm}
    \centering
    \subfigure[Degree]{
    \begin{minipage}[t]{0.48\linewidth}
    \centering
    \includegraphics[width=1.65in]{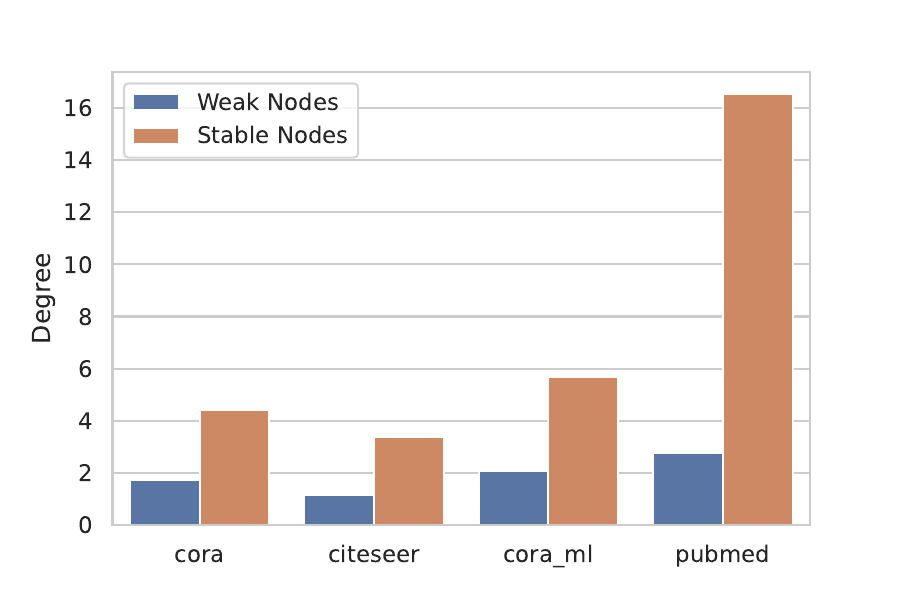}
    \end{minipage}
    }%
    \subfigure[Classification Margin]{
    \begin{minipage}[t]{0.48\linewidth}
    \centering
    \includegraphics[width=1.65in]{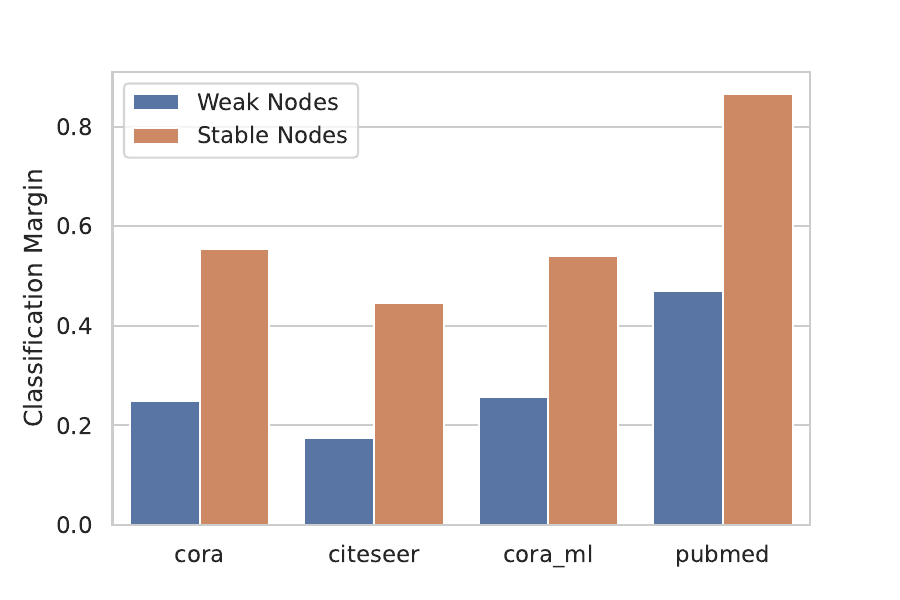}
    \end{minipage}
    }%
    \centering
    \vspace{1ex}
    \caption{Statistics of degrees and classification margins of weak nodes and stable nodes. 
    Weak nodes denote items that can be attacked successfully by GreedyRBCD. 
    Stable nodes indicate those items whose classification results do not change before and after the attack.}
    \vspace{-2ex}
    \label{degree cmp}
\end{figure}

After the preprocessing phase, we obtain node set $\mathcal{P}_{flt}(V)$ that contains nodes considered to be correctly classified by the model. 
In $\mathcal{P}_{flt}(V)$, some nodes are more stable and can be correctly classified despite a small number of fake neighbors, while others are more fragile and can be easily affected by fake edges added by attackers. 
According to the previous analysis, the two statistics, i.e., node degree and classification margin, are closely correlated with the likelihood of a successful attack on a node. 
And we focus on the combination of degree and classification margin as determining factors. 

Additionally, we employ the GreedyRBCD attack to calculate the degree and classification margin for the two types of nodes. Predictably, we find in Figure~\ref{degree cmp}(a) that the node degree has an important effect on node robustness. 
Nodes with large degrees tend to be capable of resisting several spuriously connected edges. This observation is consistent with the analysis in \cite{RBCDAttack,jin2021adversarial}.
Thus, we can first screen out the nodes with larger degrees, as they may be more stable, and attacking them requires adding more spurious edges. If we attack stable nodes, on the one hand, we need more attack budgets and on the other hand, the success rate of the attack may be lower. The degree filter can be defined as follows: 
  \begin{equation}
    \mathcal{D}_{flt}(V) = \{ v | \text{Degree}(v) < \text{threshold}_d ,~v \in \mathcal{P}_{flt}(V) \}
  \end{equation} 
where $\text{threshold}_d$ is a hyperparameter. 

\subsubsection{Margin Filter}
Furthermore, according to Figure~\ref{degree cmp}(b), the classification margin also differs significantly between the weak nodes and stable nodes. Nodes with larger classification margins tend to maintain their classification results under attack, 
while those with smaller margins are more vulnerable to fake neighbors. Thus, we further propose the margin filter to select the attack target nodes. Specifically,
\begin{equation}
  \mathcal{M}_{flt}(V) = \{ v | \text{CM}(v) < \text{threshold}_m ,~v \in \mathcal{P}_{flt}(V) \}
\end{equation} where $\text{threshold}_m$ is set the same with $\text{threshold}_d$ in the experiments.

After the proposed three filters, already misclassified nodes and the nodes with large degrees and classification margins are screened out, leaving relatively easy-to-attack nodes to become our partial attack targets, i.e., Attack($V$).

\begin{equation}
  \text{Attack}(V) = \{ v| v \in \mathcal{D}_{flt}(V) \cap \mathcal{M}_{flt}(V) \} \label{select}
\end{equation}

\subsection{Cost-effective Anchor-picking Policy}
\label{ss: cost}
The following step is to create fake connections for attack targets or to remove some edges that already exist in the graph. In the aggregation paradigm, for a node, the creation of a new connection with any other node may affect its final classification result. Here, we use \emph{anchors} to denote the nodes connected by to-be-added fake edges with attack targets and the nodes connected by to-be-removed existing edges. 

If we try to explore every potential connection relationship for adding or removing, the overhead of time and space can be unbearable. For example, Pubmed~\cite{sen2008collective} is a commonly used graph dataset for adversarial attacks, and around 20 GB of memory is required for global attacks based on its dense adjacency ~\cite{RBCDAttack}, and around 35 minutes is needed to perform a global attack with the commonly used attacker PGD~\cite{PGDAttack}. 

To reduce the attack overhead, we propose a cost-effective anchor-picking policy to select the anchors, without full exploration. Inspired by \cite{SGA}, only anchors that have the best chance of influencing attack targets are considered.

\textbf{Anchors for adding fake edges}. 
The second largest of the model prediction logits can be viewed as the second prediction category. \cite{Nettack} pointed out that if adding fake edges with attack targets, the nodes in the second prediction category are more possible to influence the classification results of attack targets. For the attack target $v$, the second largest model prediction component of the logit is $\mathbf{\tilde{z}}_{v}^{1}$.

Let $\mathcal{S}^v$ be the set of nodes in the second prediction category, which is  formulated as follows:

\begin{equation}
    \mathcal{S}(v) = \{u \ | \  \mathbf{\tilde{z}}_{u}^{0} = \mathbf{\tilde{z}}_{v}^{1}, \  u \in V \}
\end{equation}

Then, we construct the initial anchor's pool:
\begin{equation}
    \mathcal{C}_{\text{add}}(V) = \{ u \ | \ u \in \cup_{v \in \text{Attack}(V)} \mathcal{S}(v), u \in V \} 
\end{equation}
Next, we can define the pool of fake edges as $E_{\text{fake}}$. $(u,v)$ and $(v,u)$ are regarded as the same edge. 
\begin{equation}
    E_{\text{fake}} = \{  e | e = (u,v),  u \in \mathcal{C}_{\text{add}}(V) \wedge\ v \in \text{Attack}(V) \}
\end{equation}

In general, the size of $E_{\text{fake}}$ is huge. We further employ a surrogate model to reduce the size of the anchor's pool. The detailed steps are as follows: 
\begin{itemize}
    \item [$\bullet$] Calculate the edge gradient $\nabla_{e}$ for each item in $E_{\text{fake}}$.
    \item [$\bullet$] Sort them according to the edge gradient $\nabla_{e}$, and pick out the edges $E'_{\text{fake}}$ with the top-$k$ largest gradients.
        % \begin{equation}
        %     E^1_{\text{fake}} = \{ e | \nabla_{e} \text{ is the top-k largest}, e \in E_{\text{fake}} \}
        % \end{equation}
    \item [$\bullet$] Take out the nodes connected by edges in $E'_{\text{fake}}$ as the final anchor's pool:
        \begin{equation}
            \mathcal{C}_{\text{add}}(V) = \{ u | (u,v) \in E'_{\text{fake}} \} \label{anchoradd}
        \end{equation} 
\end{itemize}

The details of the surrogate model and the edge gradient $\nabla_{e}$ will be introduced in Section~\ref{ss:iterative}. \\%
\textbf{Anchors for removing fake edges}.  
Due to the over-smoothing problem, most GNN models only aggregate $k$-hop neighbors' information.
As a result, we just pick the $k$-hop neighbors of the attack targets as anchors for removing edges. 
Let $\text{Dis}(u, v)$ denote the shortest distance between $u$ and $v$. 
  \begin{equation}
    \mathcal{C}_{\text{rem}}(V) = \{ u \ | \ \text{Dis}(u, v) \leq k, ~v \in \text{Attack}(V) \} \label{anchorremove}
  \end{equation} 

With the cost-effective anchor-picking policy, we can select the anchors that have the best chance of destroying the classification results of the attack targets. Next, we propose an iterative greedy-based attack method to construct the adversarial graph by adding and removing the edge between anchors and the attack targets.

\subsection{Iterative Greedy-Based Attack}
\label{ss:iterative}
We first introduce the iterative greedy-based attack module in detail. 
Since the attacker has no access to the parameters of the target model under the setting of the white box attack, we need to train a surrogate model. 
The surrogate model can help us to compute edge gradients to decide which edge to flip in each iteration. 
Besides, a suitable attack loss should be employed. 
Following the common attack budget constraint, it is guaranteed that no more than a fixed percentage of
edges are added or removed, e.g. 1\%, 3\%, 5\%. 
We take the number of edges that can be modified as the number of iterations. 
In each iteration $t$, we greedily pick the edge with the largest edge gradient to flip, get a new perturbed graph $A^{t+1}$ (i.e., adjacency matrix), and iterate on until the attack budget is exhausted.
Two sets of candidate edges are as follows:
\begin{eqnarray}
    \mathcal{E}_{\text{add}} &=& \{ e | e=(u,v),  u \in \mathcal{C}_{\text{add}}(V) \wedge v \in \text{Att}(V)  \} \\
    \mathcal{E}_{\text{rem}} &=& \{ e |e=(u,v), u \in \mathcal{C}_{\text{rem}}(V) \wedge v \in \text{Att}(V) \} 
\end{eqnarray} 

% Each item in two sets is generated based on the results of Section~\ref{ss: hierarchical} and Section~\ref{ss: cost}. 
$\mathcal{E}_{\text{add}}$ contains edges to-be-added, and $\mathcal{E}_{\text{remove}}$ contains edges to-be-removed.
Next, we introduce the surrogate model, attack loss, edge gradient, and iterative greedy-based attack, separately. To accelerate the attack, we further propose an optimization technique with a more aggressive edge flip.

\subsubsection{Surrogate Model}
In many adversarial attack methods, a common approach is to employ a surrogate model that emulates the gradient information of the target model \cite{Nettack, RBCDAttack, Mettack}. This surrogate model is utilized to guide the attack by determining which edges should be added or deleted. Similarly, in alignment with the majority of these works, we adopt a surrogate model known as a 2-layer vanilla Graph Convolutional Network (GCN) with non-linear transformations removed, which can be formulated \cite{GCN} as follows: 
\begin{equation}
  \mathbf{Z} = f_{\theta}(A, X) = \text{softmax}(\hat{A} (\hat{A}XW^{(0)})W^{(1)} )
\end{equation} where $\hat{A}=\tilde{D}^{-\frac{1}{2}} \tilde{A} \tilde{D}^{-\frac{1}{2}}, ~\tilde{A}=A+I_{N}~\text{and}~\tilde{D}_{ii}=\sum_{j}\tilde{A}_{ij}$.
\subsubsection{Attack Loss}
Next, we describe the attack loss used by the surrogate model.
The commonly used attack losses are negative CE (Cross-Entropy) loss and CW (Carlini-Wagner) loss \cite{carlini2017towards, ying2018graph}.
However, \cite{RBCDAttack} pointed out that these two losses are flawed, with a large gap between the attack loss and the real attack effect. Attackers pay attention to already misclassified nodes, resulting in a decrease in the loss but without a decrease in the classification accuracy. In this work, we use the combination of \emph{tanhMargin} and \emph{masked CE} as the attack losses, which is proposed in \cite{RBCDAttack}. 
The loss is formulated as follows: 
\begin{eqnarray}
  \mathcal{L}_{attack} &=& 
  \sum_{v \in \text{Attack}(V)} \tanh (\tilde{z}^1_v - \tilde{z}^0_v) \\ \nonumber
  &~& + 
  \frac{1}{\| \mathbb{V}^{+} \|} \sum_{v \in \mathbb{V}^{+}} - \log \tilde{z}^0_v, 
\end{eqnarray} where $\mathbb{V}^{+}$ denotes the set of correctly classified nodes. Since we have no access to the true labels of $V_U$, we consider a node being attacked successfully as long as the predicted value of the target is not the same as the initially predicted label.

\subsubsection{Edge Gradient} 
Given a surrogate model and a suitable attack loss, the key problem is how to choose which item in $\{\mathcal{E}_{\text{add}} \cup \mathcal{E}_{\text{rem}} \}$ to modify. For continuous variables, the gradient can be calculated once per iteration, updating the variable until the loss value is sufficiently small \cite{FGSM}. However, each item in $\{\mathcal{E}_{\text{add}} \cup \mathcal{E}_{\text{remove}} \}$ is discrete, which cannot be updated directly \cite{Nettack}. PGD \cite{PGDAttack} views the edges as continuous variables (i.e., elements in the adjacency matrix), updates them iteratively with gradients, and finally samples them back into discrete variables. In this work, we also compute the gradient of each candidate edge and use the edge gradient to assist in the generation of the adversarial graph. Differently, we only calculate the gradient once and use the resulting gradient as a score \cite{Nettack, SGA}. Then, we decide which edges to flip according to this score. The computation of edge gradients in the $t$-th iteration can be formulated as follows:
\begin{equation}
    \nabla_{e}
    = \frac{\partial \mathcal{L}}{\partial e} 
    = \frac{\partial \mathcal{L}}{\partial \mathbf{Z}} \frac{\partial \mathbf{Z}}{\partial e} 
    = \frac{\partial \mathcal{L}}{\partial f_{\theta}(A, X)} \frac{\partial f_{\theta}(A, X)}{\partial e} 
    \label{calcg}
\end{equation} 

Note that the computation of the attack loss $\mathcal{L}$ only considers the attack targets selected by the hierarchical target selection policy. The smaller the attack loss is, the better performance the attack can get. The gradient value measures how fast the loss function changed about that variable. A larger edge gradient indicates that this edge is more effective in reducing attack losses.

\subsubsection{Iterative Greedy-Based Attack}
Given the gradient values for all edges in $\mathcal{E}_{\text{add}}$ and $\mathcal{E}_{\text{remove}}$, we greedily flip the edge with the largest gradient value in each iteration $t$.
%which is the local optimum operation at the current moment. 
After that, we remove this edge from $\mathcal{E}_{\text{add}}$ or $\mathcal{E}_{\text{remove}}$, and add it to $A^t$:
\begin{eqnarray}
    e_* &=& \arg\max_{e} \{\nabla_{e} \ | \ e \in \mathcal{E}_{\text{add}} \cup \mathcal{E}_{\text{remove}} \} \\ 
    A^{t+1} &=& \text{Union}(A^t, e_*) \label{modifiedA}
\end{eqnarray} where $\text{Union}(A, e)$ denotes adding an edge $e$ to the adjacency matrix $A$ if $e \in \mathcal{E}_{\text{add}}$, or removing an edge $e$ from $A$ if $e \in \mathcal{E}_{\text{remove}}$.
In the next iteration, $A^{t+1}$ will be used in the surrogate model.

\subsubsection{More Aggressive Edge Flip}
When the number of edges in $\{\mathcal{E}_{\text{add}} \cup \mathcal{E}_{\text{remove}}\}$ becomes large, 
the time overhead of this greedy-based attack approach rises significantly, e.g. attacks on large graphs such as Pubmed and ogbn-arxiv. To solve this problem, we use a more aggressive edge flip method, which computes the gradient once and flips multiple $K$ candidate edges. $K$ is called $\text{greedy step}$. In practice, we find that a large number of candidate edges have the same maximum gradient value in each iteration. Due to the more aggressive edge flip, the time overhead of \mymodel{} can be significantly reduced, especially on large graphs. The workflow of \mymodel{} is described in Algorithm~\ref{algori}. 

\begin{algorithm}[t]
  \caption{Partial Graph Attack (PGA)}
  \label{algori}
  \SetAlgoLined
  \KwIn{Graph $\mathcal{G}=(V, A, X)$, Labeled nodes $V_L$, Unlabeled nodes $V_U$, Greedy step $K$, Budget $\Delta$}%输入参数
  \KwOut{The adversarial graph, $\mathcal{G}'=(V, A', X)$}%输出
  \BlankLine
  Normally train the victim model\;
  Train the surrogate model using the same data as the victim model\;
  Get the prediction logits of the surrogate model\;
  Select the attack targets $\text{Attack}(V)$ using Eq. \ref{select} \;
  Pick two anchor pools $\mathcal{C}_{\text{add}}$ and $\mathcal{C}_{\text{remove}}$ using Eq. \ref{anchoradd} and Eq. \ref{anchorremove} \;
  Construct the candidate edge sets $\mathcal{E}_{\text{add}}$ and $\mathcal{E}_{\text{remove}}$ \;
  \While{$\Delta$ not exhausted}{
    Compute edge gradient for each candidate edge \;
    Select the edges with the $K$ largest gradients to flip, and update the perturbed graph $A^t$ to $A^{t+1}$ using Eq. \ref{modifiedA} \; 
  }
  return $\mathcal{G}'$ \;
\end{algorithm}

\section{Experiments}
In this section, we conduct extensive experiments to answer the following questions:
\begin{itemize}
 \item {\bfseries RQ1}: How is the effectiveness of the proposed \mymodel{} compared with existing white box graph  attack methods?
  \item {\bfseries RQ2}: How is the  attack effect under different attack budgets? 
 \item {\bfseries RQ3}: How about the  attack efficiency of \mymodel{}? 
 \item {\bfseries RQ4}: What is the impact of each major component of \mymodel{}?
 \item {\bfseries RQ5}: Is the perturbation graph generated by \mymodel{} unnoticeable?
  \item {\bfseries RQ6}: How is the effectiveness when transferring \mymodel{} to the poisoning attack setting?
 \item {\bfseries RQ7}: what are the influences of different hyper-parameter values?
\end{itemize}
%

% Our experimental setup comprises several components. Firstly, we provide an overview of the fundamental settings, including details about the datasets used, the victim model employed, the baselines utilized, and the implementation details. Following this, we delve into the primary experiment, which involves an adversarial attack with a 5 percent perturbation. This experiment is divided into two categories: one that involves models without any defense mechanisms, and the other that incorporates various defense strategies. While our primary focus lies on evasion attacks, we also conducted some preliminary experiments regarding poisoning attacks. For these experiments, we utilized the perturbed graph obtained through the evasion attack and employed it directly for the poisoning attack test.
% %

% Next, we proceed to conduct attack experiments ranging from 1 percent to 5 percent perturbation. The purpose of these experiments is to compare the attack performance of different baselines under varying levels of perturbations. Subsequently, we analyze the time complexity of \mymodel{}. 
% %

% Lastly, we perform ablation experiments. Additionally, we conduct hyperparameter experiments to explore the effects of different parameter values on the overall performance of our attack method. These experiments allow us to gain a deeper understanding of the robustness and effectiveness of our approach.
% %

\subsection{Experimental Settings}
\begin{table}[h] 
  \centering
  \caption{Statistics of the datasets.}
  \begin{tabular}{c|cccc}
    \toprule
    & \textbf{\#Nodes} & \textbf{\#Edges} & \textbf{\#Features} & \textbf{\#Classes} \\
    \midrule
    \textbf{Cora} & 2708 & 5278 & 1433 & 7 \\
    \textbf{Citeseer} & 3327 & 4552 & 3703 & 6 \\
    \textbf{CoraML} & 2995 & 16316 & 2879 & 7 \\
    \textbf{Pubmed} & 19717 & 44324 & 500 & 3 \\
    \bottomrule
  \end{tabular}
  \label{experiments: dataset}
\end{table}
\subsubsection{Datasets}
We evaluate PGA on four real-world benchmark datasets (Cora, Citeseer, CoraML, and Pubmed).
The statistics of these datasets are shown in Table \ref{experiments: dataset}.

\subsubsection{Victim Models}
We choose a comprehensive set of ten GNN models as victim models, which can be divided into two groups: five GNN models without defense mechanisms and five defense models equipped with various defense strategies. 
Among the GNN models without defense mechanisms, we select widely used GNNs such as the Vanilla GCN \cite{GCN}, GAT \cite{velivckovic2017graph}, SGC \cite{SGC}, GraphSAGE \cite{hamilton2017inductive}, and APPNP \cite{APPNP}.
\begin{itemize}
    \item [$\bullet$] GCN \cite{GCN}. GCN is one of the most common GNNs that learn node representations by aggregating neighbors. 
    \item [$\bullet$] GAT \cite{velivckovic2017graph}. GAT uses an attention mechanism to learn the importance of different neighbors. 
    \item [$\bullet$] SGC \cite{SGC}. SGC removes the linear transformation of the GCN, improving model training efficiency while keeping performance largely unchanged.
    \item [$\bullet$] GraphSAGE \cite{hamilton2017inductive}. GraphSAGE is an inductive framework that samples neighbors for node aggregation.
    \item [$\bullet$] APPNP \cite{APPNP}. PPNP introduces a personalized PageRank-based aggregation operation and employs an MLP for feature transformation, while APPNP is an approximated version of PPNP that balances computational efficiency and effectiveness in capturing personalized information.
\end{itemize}

Moreover, the defense models encompass a range of different strategies. 
Specifically, we incorporate RGCN \cite{RGCN}, MedianGCN \cite{chen2021understanding}, GNNGuard \cite{GNNGuard}, Jaccard \cite{Jaccard}, and Grand \cite{Grand}.  Note that the selection of these five defense models is not arbitrary. 
Instead, we deliberately choose defense models that represent distinct defense strategies and are currently among the most popular choices in the field.

\begin{itemize}
    \item [$\bullet$] RGCN \cite{RGCN}. RGCN uses Gaussian distribution in hidden layers and variance-based attention weights to enhance the robustness of GCN against adversarial attacks. %which is a novel graph convolution model.
    \item [$\bullet$] MedianGCN \cite{chen2021understanding}. MedianGCN uses more robust aggregation functions such as trimmed mean and median to defend against adversarial attacks.
    \item [$\bullet$] GNNGuard \cite{GNNGuard}. GNNGuard utilizes cosine similarity for computing defense coefficients during each aggregation layer, assigning greater importance to similar neighbor information while obstructing abnormal neighbor information. To enhance stability, the current defense coefficients retain a partial memory of the preceding layer.
    \item [$\bullet$] Jaccard \cite{Jaccard}. Jaccard employs a straightforward and efficient approach by incorporating a threshold to directly discard neighboring information with low similarity.
    \item [$\bullet$] Grand \cite{Grand}. Grand relies on random feature augmentations (zeroing features) coupled with neighborhood augmentations $\bar{X} = (AX + AAX + \dots)$ \cite{are_robust_nips22}. All randomly augmented copies of $\bar{X}$ are passed through the same MLP that is trained with a consistency regularization loss \cite{are_robust_nips22}.
\end{itemize}
\subsubsection{Baselines}
While numerous research studies have explored graph adversarial attacks, not all of them are capable of conducting fair comparisons. In this study, we have carefully chosen the following baselines that specifically concentrate on evasion attacks under the white box settings.
(1) \textbf{Random}: randomly perturb the graph data. 
(2) \textbf{DICE} \cite{waniek2018hiding}: delete edges internally, and connect edges externally. 
(3) \textbf{Greedy} \cite{Mettack}: a variant of meta-self attack without weight re-training. 
(4) \textbf{tanh-PGD} \cite{PGDAttack}: introduce an edge perturbation-based topology attack framework from a first-order optimization perspective. We change the attack loss to tanhMargin \cite{RBCDAttack}.
(5) \textbf{CW-PGD} \cite{PGDAttack}: belong to PGD but use the CW attack loss. 
(6) \textbf{PRBCD} \cite{RBCDAttack}: improve PGD by combining random block gradient descent.
(7) \textbf{GreedyRBCD} \cite{RBCDAttack}: an alternative to PRBCD that incorporates the greedy strategy.

\subsubsection{Implementation details}
To initiate the process, we begin by training the victim models. For instance, when evaluating GCN on the Cora dataset, we employ a 2-layer GCN with ReLU nonlinear transformation. The training process involves using the Adam optimizer for 200 epochs, setting a patience of 30, a hidden layer dimension of 16, and a dropout rate of 0.5. These values correspond to the default GCN hyperparameters of Deeprobust \cite{li2020deeprobust}, which is the most popular third-party graph attack library. 
During the training of victim models, the objective is to maximize the accuracy of the model on the clean dataset. 
Certain defense models may require an extended training setting. For example, GNNGuard~\cite{GNNGuard} undergoes 3000 epochs of training, with patience set to 100. The dimension of the hidden layer remains fixed at 32, and the dropout rate is set to 0.5. We refer to the original author's code for specific model-specific hyperparameters. 
Furthermore, the learning rate and weight decay of the Adam optimizer are fixed at 0.01 and 0.0005, respectively, following the settings outlined in Deeprobust.

Subsequently, we introduce the details of the graph attack methods, which are designed to disrupt the performance of the victim models. We implement PRBCD and GreedyRBCD using the code provided by the original papers, ensuring that we maintain consistency with their implementations. For the remaining baselines, we utilize Deeprobust~\cite{li2020deeprobust}. The hyperparameters for these baselines are also set in accordance with the settings specified by the original authors.

% Except for PRBCD and GreedyRBCD \cite{RBCDAttack}, all other baselines are implemented using the third-party library Deeprobust~\cite{li2020deeprobust}. For PRBCD and GreedyRBCD, we use the author's code \cite{geisler2021robustness}. When training the victim models, we choose the best hyperparameters possible to achieve the best classification accuracy. Then, the architectures and parameters of the victim models remain unchanged.

\begin{table*}[t]
  \centering
  \caption{Evasion attack effect on regular models (the lower, the better) under 5\% perturbed edges. Bold and underline indicate the best and the second best, respectively. }
  \resizebox{\textwidth}{!}{
         \begin{tabular}{c|c|ccccccccc}
                \toprule
                \textbf{Victim}     & \textbf{Dataset}  & \textbf{Clean} & \textbf{Random} & \textbf{DICE}  & \textbf{Greedy} & \textbf{tanh-PGD} & \textbf{CW-PGD} & \textbf{PRBCD}                          & \textbf{GreedyRBCD}                         & \textbf{PGA}                                   \\
                \midrule
                \multirow{4}{*}{\textbf{GCN}} 
                                     & \textbf{Cora}     
                                        & 82.02±0.52  % clean
                                        & 81.20±0.50  % random
                                        & 81.40±0.48  % dice
                                        & 75.30±1.03  % greedy
                                        & 72.26±0.99  % tanh-PGD
                                        & 75.78±0.87  % cw-PGD
                                        & 70.26±0.81  % prbcd
                                        & \underline{68.90±0.42}  % grbcd
                                        & \textbf{66.26±0.59} \\  % pga
                                        
                                     & \textbf{Citeseer} 
                                         & 71.46±0.35  % clean
                                         & 71.02±0.21  % random
                                         & 70.58±0.32  % dice
                                         & 65.80±0.91  % greedy
                                         & 65.58±0.84  % tanh-PGD
                                         & 66.08±0.29  % cw-PGD
                                         & 63.64±0.71  % prbcd
                                         & \underline{62.86±1.27}    % grbcd  
                                         & \textbf{61.12±0.97}   \\  % pga
                                     
                                     & \textbf{CoraML} 
                                         & 82.94±0.55  % clean
                                         & 82.28±0.43  % random
                                         & 82.24±0.45  % dice
                                         & 74.30±0.34  % greedy
                                         & 71.68±1.14  % tanh-PGD
                                         & 74.04±0.96  % cw-PGD
                                         & 69.18±1.01  % prbcd
                                         & \underline{67.22±0.63}     % grbcd  
                                         & \textbf{62.62±0.29}    \\  % pga
                                         
                                     & \textbf{Pubmed}   
                                         & 79.18±0.37    % clean
                                         & 78.36±0.64   % random
                                         & 78.32±0.29   % dice
                                         & 58.22±0.73   % greedy
                                         & 68.92±1.33   % tanh-PGD
                                         & 54.64±0.92   % cw-PGD
                                         & 41.10±1.06   % prbcd
                                         & \textbf{29.58±0.43}         % grbcd 
                                         & \underline{29.98±0.40}  \\  % pga                                   
                                     \midrule
                \multirow{4}{*}{\textbf{GAT}} 
                                     & \textbf{Cora}     
                                        & 83.12±0.49 
                                        & 82.34±0.81  
                                        & 82.62±0.56  % dice
                                        & 76.50±0.72  
                                        & 74.48±1.19  % tanh-PGD
                                        & 76.46±0.83  
                                        & 72.52±1.20  % prbcd                       
                                        & \underline{72.02±0.91}      
                                        & \textbf{71.22±0.42}              \\
                                        
                                     & \textbf{Citeseer} 
                                        & 71.38±0.86 
                                        & 70.62±0.71  
                                        & 70.42±0.54  % dice
                                        & 65.48±0.65  
                                        & 66.46±1.11  % tanh-PGD
                                        & 66.46±0.89  
                                        & 64.62±0.65
                                        & \underline{64.28±0.76}
                                        & \textbf{62.86±0.65}              \\
                                        
                                     & \textbf{CoraML} 
                                        & 82.90±0.50 
                                        & 81.68±0.65  
                                        & 81.36±0.64  % dice
                                        & 74.66±0.67  
                                        & 73.82±0.84  % tanh-PGD
                                        & 74.56±0.87  
                                        & 72.16±0.71                          
                                        & \underline{70.52±1.17}          
                                        & \textbf{70.16±1.08}          \\
                                        
                                     & \textbf{Pubmed}   
                                        & 78.24±0.63 
                                        & 76.80±0.73  
                                        & 77.06±0.92   % dice
                                        & 58.56±0.55  
                                        & 68.86±1.34   % tanh-PGD
                                        & 57.60±1.50  
                                        & 44.72±1.73                          
                                        & \underline{40.62±1.54}      
                                        & \textbf{37.36±1.40}              \\
                                
                                     \midrule
                \multirow{4}{*}{\textbf{SGC}} 
                                     & \textbf{Cora}     
                                        & 80.74±0.10 
                                        & 80.14±0.77  
                                        & 79.88±0.41  % dice
                                        & 73.74±0.67  
                                        & 71.74±0.77  % tanh-PGD
                                        & 73.48±0.60  
                                        & 70.96±0.55  % prbcd                         
                                        & \underline{68.68±0.75}      
                                        & \textbf{67.50±0.21}              \\
                                        
                                     & \textbf{Citeseer} 
                                        & 71.56±0.05 
                                        & 70.88±0.58  
                                        & 70.60±0.35  % dice
                                        & 65.98±0.23  
                                        & 66.18±0.99  % tanh-PGD
                                        & 66.08±0.32  
                                        & 64.82±0.80
                                        & \underline{64.34±0.85}
                                        & \textbf{62.46±1.15}              \\
                                        
                                     & \textbf{CoraML} 
                                        & 80.54±0.10 
                                        & 79.60±0.26 
                                        & 78.70±0.45  % dice
                                        & 72.74±0.43  
                                        & 70.56±0.89  % tanh-PGD
                                        & 71.16±0.69  
                                        & 69.36±1.05  
                                        & \underline{67.36±1.29}  
                                        & \textbf{64.24±0.62}              \\
                                        
                                     & \textbf{Pubmed}   
                                        & 77.40±0.00 
                                        & 76.82±0.53  
                                        & 76.78±0.20   % dice
                                        & 57.82±0.81  
                                        & 68.00±1.21   % tanh-PGD
                                        & 56.28±0.67  
                                        & 44.44±1.51                          
                                        & \underline{38.08±0.72}      
                                        & \textbf{35.58±1.38}              \\
                                        
                                     \midrule
                                     
                \multirow{4}{*}{\textbf{GraphSAGE}} 
                                     & \textbf{Cora}     
                                        & 81.52±0.33 
                                        & 80.78±0.45  
                                        & 80.90±0.21  % dice
                                        & 75.48±0.53  
                                        & 74.12±0.73  % tanh-PGD
                                        & 75.76±0.77  
                                        & 72.40±1.35
                                        & \underline{72.12±0.38}      
                                        & \textbf{71.42±0.67}              \\
                                        
                                     & \textbf{Citeseer} 
                                        & 71.76±0.26 
                                        & 71.28±0.46  
                                        & 70.92±0.35  % dice
                                        & 67.36±0.35  
                                        & 67.44±0.46  % tanh-PGD
                                        & 67.48±0.51  
                                        & 65.98±0.52                          
                                        & \underline{65.46±0.68} 
                                        & \textbf{64.12±0.60}              \\
                                        
                                     & \textbf{CoraML} 
                                        & 81.16±0.50 
                                        & 80.92±0.65  
                                        & 80.40±0.33  % dice
                                        & 74.90±0.32  
                                        & 72.66±0.48  % tanh-PGD
                                        & 74.94±1.01  
                                        & 71.74±0.63      
                                        & \underline{72.32±0.65}
                                        & \textbf{69.42±0.66}          \\
                                        
                                     & \textbf{Pubmed}   
                                        & 77.64±0.25 
                                        & 76.92±0.27  
                                        & 76.80±0.42  % dice
                                        & 58.50±0.83  
                                        & 69.58±1.52  % tanh-PGD
                                        & 62.78±0.23  
                                        & 50.38±0.75                          
                                        & \underline{48.24±0.89}      
                                        & \textbf{47.68±2.21}              \\
                    \midrule
                    \multirow{4}{*}{\textbf{APPNP}} 
                                     & \textbf{Cora}     
                                        & 83.92±0.31 
                                        & 83.40±0.50  
                                        & 83.22±0.35  % dice
                                        & 77.32±1.21  
                                        & 76.22±1.04  % tanh-PGD
                                        & 79.64±0.53  
                                        & \underline{75.14±0.89}
                                        & 75.88±0.85      
                                        & \textbf{74.60±0.64}              \\
                                        
                                     & \textbf{Citeseer} 
                                        & 72.42±0.21 
                                        & 71.44±0.26  
                                        & 70.90±0.53  % dice
                                        & 66.50±0.36  
                                        & 67.10±0.71  % tanh-PGD
                                        & 67.40±0.32  
                                        & 65.20±0.30                          
                                        & \underline{65.02±1.13} 
                                        & \textbf{64.14±0.96}              \\
                                        
                                     & \textbf{CoraML} 
                                        & 84.72±0.32 
                                        & 84.06±0.67 
                                        & 83.68±0.37  % dice
                                        & 76.92±0.58  
                                        & 75.74±1.00  % tanh-PGD
                                        & 78.66±1.38  
                                        & \underline{74.28±1.20}      
                                        & 76.38±0.77
                                        & \textbf{71.96±0.71}          \\
                                        
                                     & \textbf{Pubmed}   
                                        & 80.24±0.14 
                                        & 79.48±0.44  
                                        & 79.56±0.38  % dice
                                        & 59.44±0.43  
                                        & 70.80±1.02  % tanh-PGD
                                        & 63.08±0.54  
                                        & 47.88±1.54                          
                                        & \underline{44.82±1.11}      
                                        & \textbf{43.12±2.84}              \\
                    
                                     \bottomrule
         \end{tabular}
  }
\label{experiments: normal main}%
\end{table*}%

\begin{table*}[t]
  \centering
  \caption{Evasion attack effect on robust models (the lower, the better) under 5\% perturbed edges.  Bold and underline indicate the best and the second best, respectively.}
  \resizebox{\textwidth}{!}{
         \begin{tabular}{c|c|cccccccc}
                \toprule
                \textbf{Victim}     & \textbf{Dataset}  & \textbf{Random} & \textbf{DICE}  & \textbf{Greedy} & \textbf{tanh-PGD} & \textbf{CW-PGD} & \textbf{PRBCD}                          & \textbf{GreedyRBCD}                         & \textbf{PGA}                                   \\
                \midrule
                \multirow{4}{*}{\textbf{RGCN}} 
                                     & \textbf{Cora}     
                &81.44±0.19
                &81.64±0.41
                &75.34±1.18
                &72.58±0.87
                &75.88±0.53
                &71.64±0.57
                &\underline{70.32±0.55}
                &\textbf{68.08±1.00} \\  
                                        
                                     & \textbf{Citeseer} 
                                                         &70.88±0.38
                &70.24±0.42
                &65.84±0.59
                &65.52±0.79
                &66.00±0.39
                &63.98±0.91
                &\underline{63.36±1.09}
                &\textbf{61.14±0.75} \\
                                     
                                     & \textbf{CoraML} 
                                                         &82.66±0.71
                &82.16±0.33
                &73.96±0.47
                &72.12±0.87
                &74.78±1.03
                &69.84±0.72
                &\underline{68.14±0.55}
                &\textbf{63.74±0.12} \\
                                         
                                     & \textbf{Pubmed}   
                                                         &78.30±0.49
                &78.16±0.22
                &58.24±0.80
                &69.02±1.49
                &56.84±0.59
                &42.98±0.98
                &\underline{34.04±1.35}
                &\textbf{31.84±0.71} \\                                 
                                     \midrule
                \multirow{4}{*}{\textbf{MedianGCN}} 
                                     & \textbf{Cora}     
                                                        &79.80±0.23
                &79.98±0.57
                &75.40±0.38
                &76.12±0.88
                &77.06±0.37
                &75.32±0.44
                &\underline{75.18±0.55}
                &\textbf{74.88±0.57} \\
                                        
                                     & \textbf{Citeseer} 
                                                        &70.86±0.87
                &70.42±0.73
                &67.78±0.92
                &67.12±0.85
                &66.96±0.47
                &\underline{66.00±0.42}
                &66.60±0.41
                &\textbf{65.82±0.62}\\
                                        
                                     & \textbf{CoraML} 
                                                        &83.48±0.52
                &83.28±0.46
                &\underline{75.38±0.47}
                &77.54±0.50
                &78.14±0.54
                &76.86±0.86
                &\textbf{74.66±0.54}
                &75.86±1.08\\
                                        
                                     & \textbf{Pubmed}  
                                                        &77.50±0.29
                &77.12±0.43
                &58.72±0.52
                &70.58±1.51
                &60.38±1.26
                &49.08±1.01
                &\underline{47.36±1.00}
                &\textbf{46.26±1.16}\\
                                
                                     \midrule
                \multirow{4}{*}{\textbf{Jaccard}} 
                                     & \textbf{Cora}  
                                                        &78.74±0.69
                &78.76±0.65
                &75.36±0.90
                &75.12±0.99
                &76.56±1.04
                &75.12±0.68
                &\underline{74.20±0.79}
                &\textbf{72.80±0.98}\\
                                        
                                     & \textbf{Citeseer} 
                                                        &70.32±0.44
                &70.16±0.61
                &68.74±0.68
                &68.08±1.03
                &68.84±0.36
                &67.88±0.48
                &\underline{67.84±0.71}
                &\textbf{67.62±0.46}\\
                                        
                                     & \textbf{CoraML} 
                                        &79.78±0.53
&79.32±0.48
&75.76±0.62
&75.46±1.20
&75.96±0.31
&\underline{74.20±0.85}
&75.76±0.63
&\textbf{71.94±0.55}\\
                                        
                                     & \textbf{Pubmed}  
                                        &78.34±0.40
&78.26±0.12
&58.76±1.06
&69.48±1.44
&59.78±0.91
&47.50±0.69
&\underline{44.04±0.77}
&\textbf{41.56±1.76}\\
                                        
                                     \midrule
                                     
                \multirow{4}{*}{\textbf{Grand}} 
                                     & \textbf{Cora} 
                                        &84.16±0.17
&84.36±0.50
&78.64±0.71
&77.54±0.23
&80.86±0.54
&\underline{76.70±0.63}
&77.64±0.74
&\textbf{76.58±0.33}\\
                                        
                                     & \textbf{Citeseer} 
                                       &73.40±0.26
&73.28±0.34
&70.02±0.89
&69.70±0.54
&69.90±0.23
&\textbf{68.62±0.84}
&69.92±1.14
&\underline{68.98±0.68}\\
                                        
                                     & \textbf{CoraML} 
                                        &84.64±0.32
&84.66±0.23
&78.60±0.70
&78.02±0.95
&80.80±0.93
&\underline{76.98±0.71}
&78.22±0.67
&\textbf{75.96±0.89}\\
                                        
                                     & \textbf{Pubmed} 
                                        &79.26±0.39
&79.24±0.31
&59.02±0.57
&71.04±1.40
&65.34±0.44
&51.70±1.17
&\underline{46.56±0.66}
&\textbf{45.60±1.97}\\
                    \midrule
                    \multirow{4}{*}{\textbf{GNNGuard}} 
                                     & \textbf{Cora}   
                                        &80.46±0.19
&80.56±0.40
&76.56±0.52
&76.66±0.46
&78.02±0.40
&\underline{76.10±0.74}
&76.34±0.42
&\textbf{76.08±0.21}\\
                                        
                                     & \textbf{Citeseer} 
                                        &70.10±0.55
&70.08±0.37
&69.06±0.30
&68.54±0.33
&68.62±0.43
&\underline{67.96±0.85}
&68.30±0.46
&\textbf{67.66±0.53}\\
                                        
                                     & \textbf{CoraML} 
                                        &80.64±0.31
&80.24±0.58
&77.22±0.61
&77.68±0.74
&78.04±0.69
&76.96±0.83
&\underline{76.96±0.54}
&\textbf{76.34±0.49}\\
                                        
                                     & \textbf{Pubmed} 
                                        &77.54±0.22
&77.68±0.49
&59.50±0.62
&70.54±1.17
&65.92±0.91
&\textbf{52.36±0.86}
&\underline{55.80±1.71}
&56.08±3.21\\
                    
                                     \bottomrule
         \end{tabular}
  }
\label{experiments-robust-main}%
\end{table*}%

\subsection{Comparison of Attack Effect (RQ1)}
%\subsubsection{5\% Perturbation}
We conduct adversarial attacks with a 5\% perturbation level. To perform these attacks, we first train the victim model.
%which is the model we aimed to attack. 
The evasion attack is then executed strictly based on the setup of a grey-box attack, meaning that we do not have access to the parameters of the victim model. Additionally, the attackers could only obtain the labels of the training data, limiting the information available for the attack. 
To evaluate the effectiveness of the attacks, we use classification accuracy as the evaluation metric. After the attacks are executed, we measure the classification accuracy of the victim model on the perturbed graph. 

Table \ref{experiments: normal main} presents the classification accuracy after the attacks. From Table \ref{experiments: normal main}, we see that the proposed \mymodel{} achieves better attack effects compared to existing attack methods in most cases. 
We draw the following conclusions regarding the performance of the attacks and the effectiveness of \mymodel{}:
\begin{itemize}
  \item[$\bullet$] The random method does not result in remarkable model performance degradation. The attack effect is particularly insignificant on models with special aggregation operations, such as GAT and APPNP. The same is true for the DICE attack, where the performance degradation of the models is hardly visible.
  \item[$\bullet$] The greedy method shows promising results and can cause performance degradation in various scenarios with different datasets and GNN models. However, its attack effect is still limited since it is a modification of the meta-attack approach described in \cite{Mettack}. Additionally, as the dataset size increases, the time overhead dramatically rises due to the consideration of all candidate edges.
  \item[$\bullet$] tanh-PGD and CW-PGD outperform the aforementioned baselines. They allow for more efficient exploration in generating adversarial graphs from an optimization standpoint. However, as pointed out in \cite{RBCDAttack}, the performance of PGD is limited by its loss function, and its time and space complexity of $\mathcal{O}(n^2)$ significantly increases on slightly larger datasets.
  \item[$\bullet$] PRBCD and GreedyRBCD prove to be highly effective attack methods, consistently performing well in all scenarios. The random block gradient descent employed by these methods contributes to their scalability. Similar to other baselines, they still attack all objects uniformly.
  \item[$\bullet$] Our proposed \mymodel{} achieves the best attack effect across different victim models and datasets, demonstrating the effectiveness of the partial graph attack that directly targets vulnerable items instead of all nodes.
\end{itemize}

The set of experiments in Table \ref{experiments: normal main} evaluate the attack performance of different attack methods on five GNN models trained without any defense mechanisms. To provide a more comprehensive evaluation of these attack methods, we conduct an additional set of attack experiments on five GNN models equipped with defense mechanisms. The results presented in Table~\ref{experiments-robust-main} also demonstrate that our proposed method maintains a strong attack performance even when applied to the GNN model with defense mechanisms.
\subsection{Attack Effect Under Different Budgets (RQ2)}

\begin{figure}[t]
  \setlength{\abovecaptionskip}{0.2cm}
  \centering
  \subfigure[Cora]{
  \hspace{-4mm}
  \begin{minipage}[t]{0.5\linewidth}
  \centering
  \includegraphics[width=1.85in]{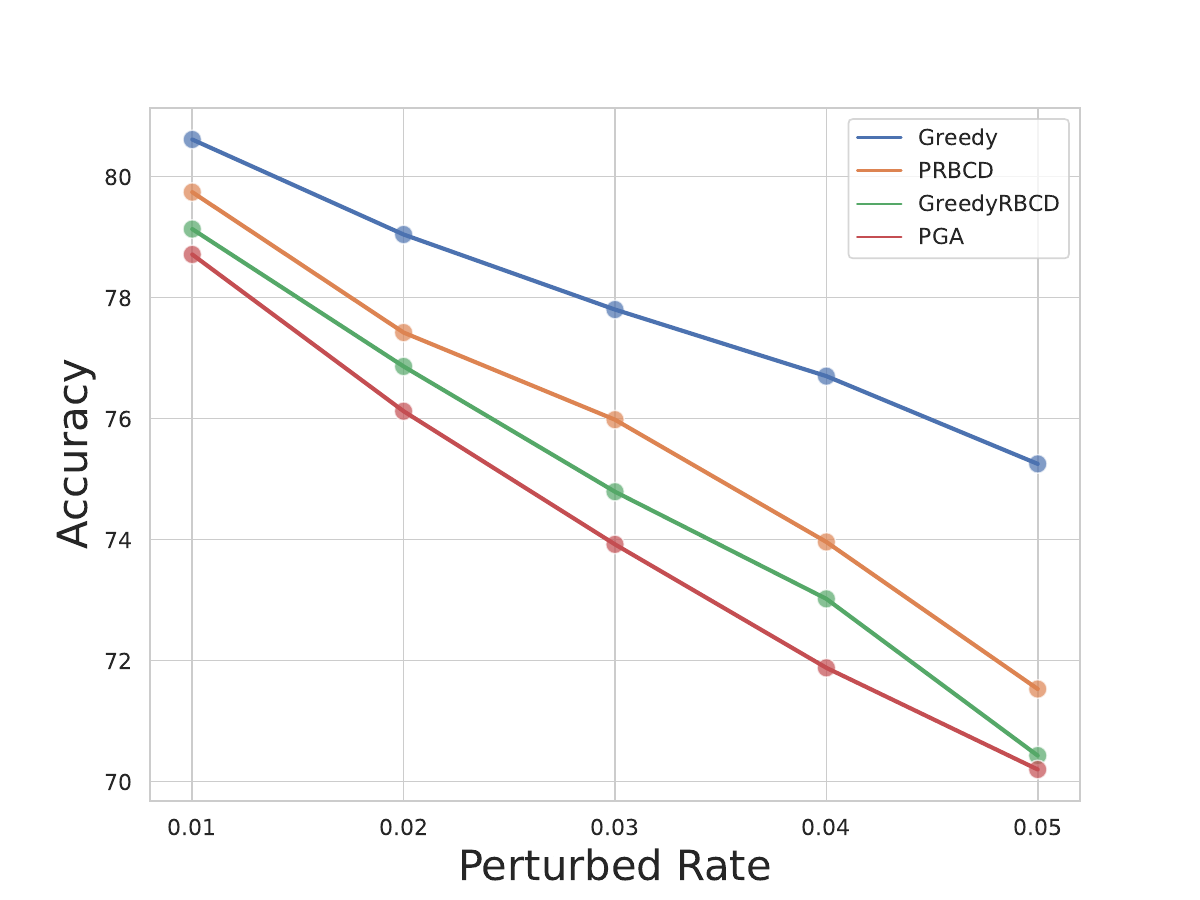}
  %\caption{fig2}
  \end{minipage}
  }%
  \subfigure[Citeseer]{
  \begin{minipage}[t]{0.5\linewidth}
  \centering
  \includegraphics[width=1.85in]{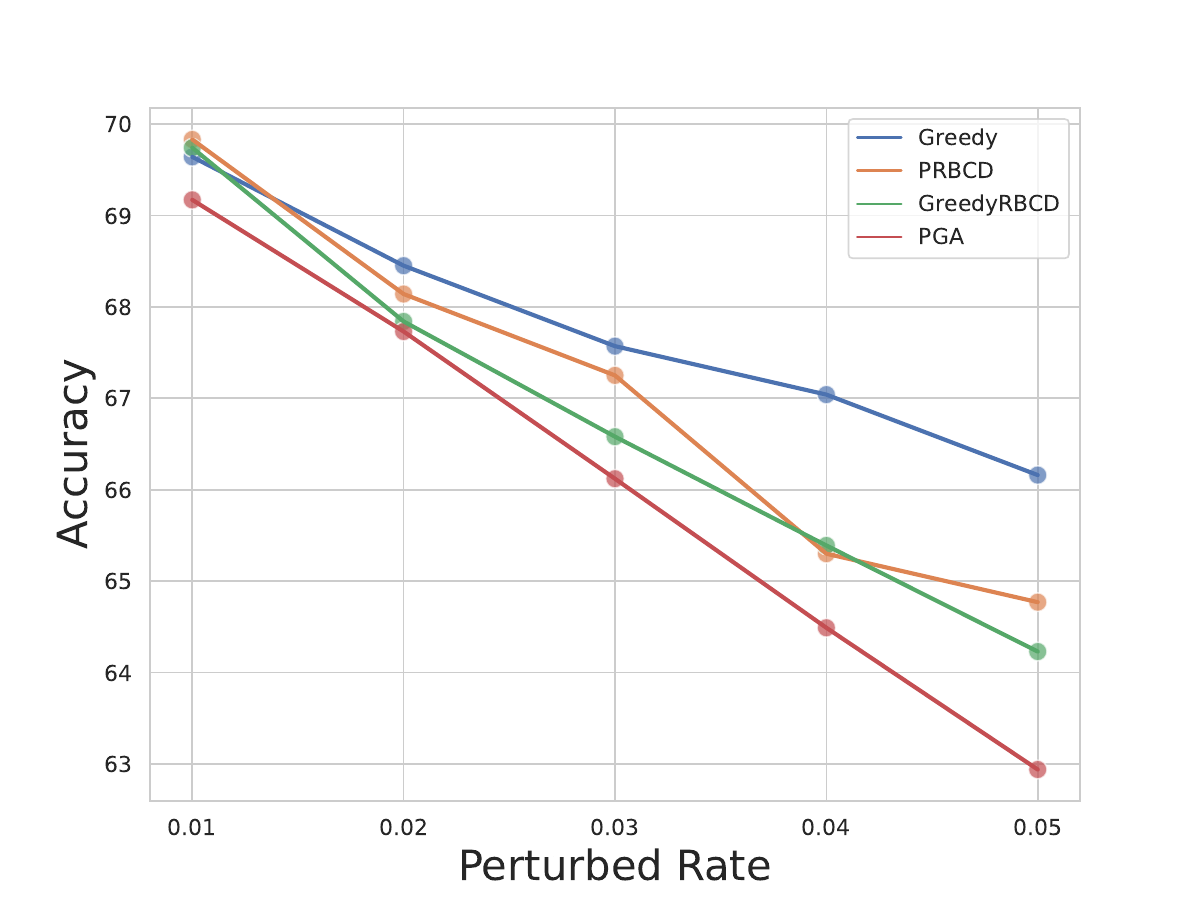}
  %\caption{fig2}
  \end{minipage}
  }%
  \quad
  \centering
  \vspace{-4mm}
  \subfigure[CoraML]{
  \hspace{-4mm}
  \begin{minipage}[t]{0.5\linewidth}
  \centering
  \includegraphics[width=1.85in]{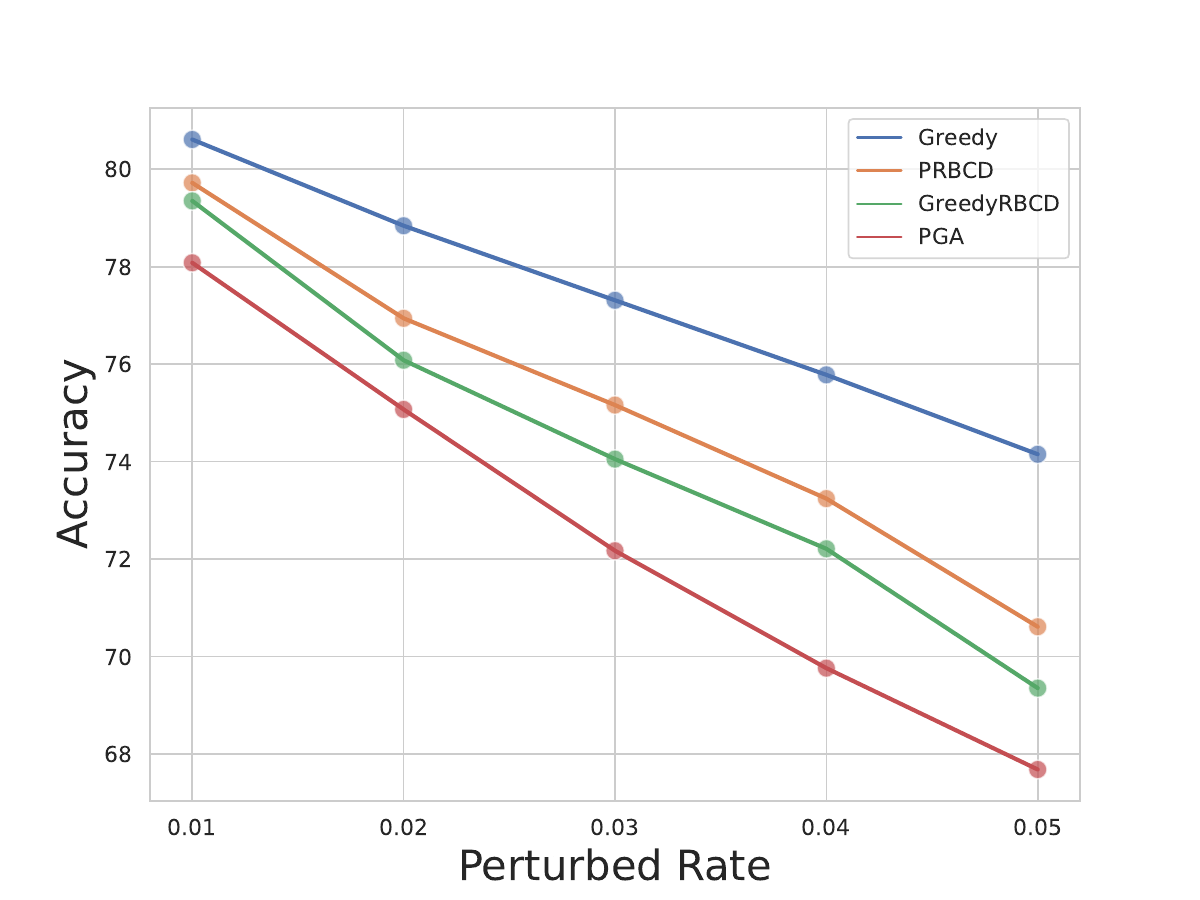}
  %\caption{fig2}
  \end{minipage}
  }%
  \subfigure[Pubmed]{
  \begin{minipage}[t]{0.5\linewidth}
  \centering
  \includegraphics[width=1.85in]{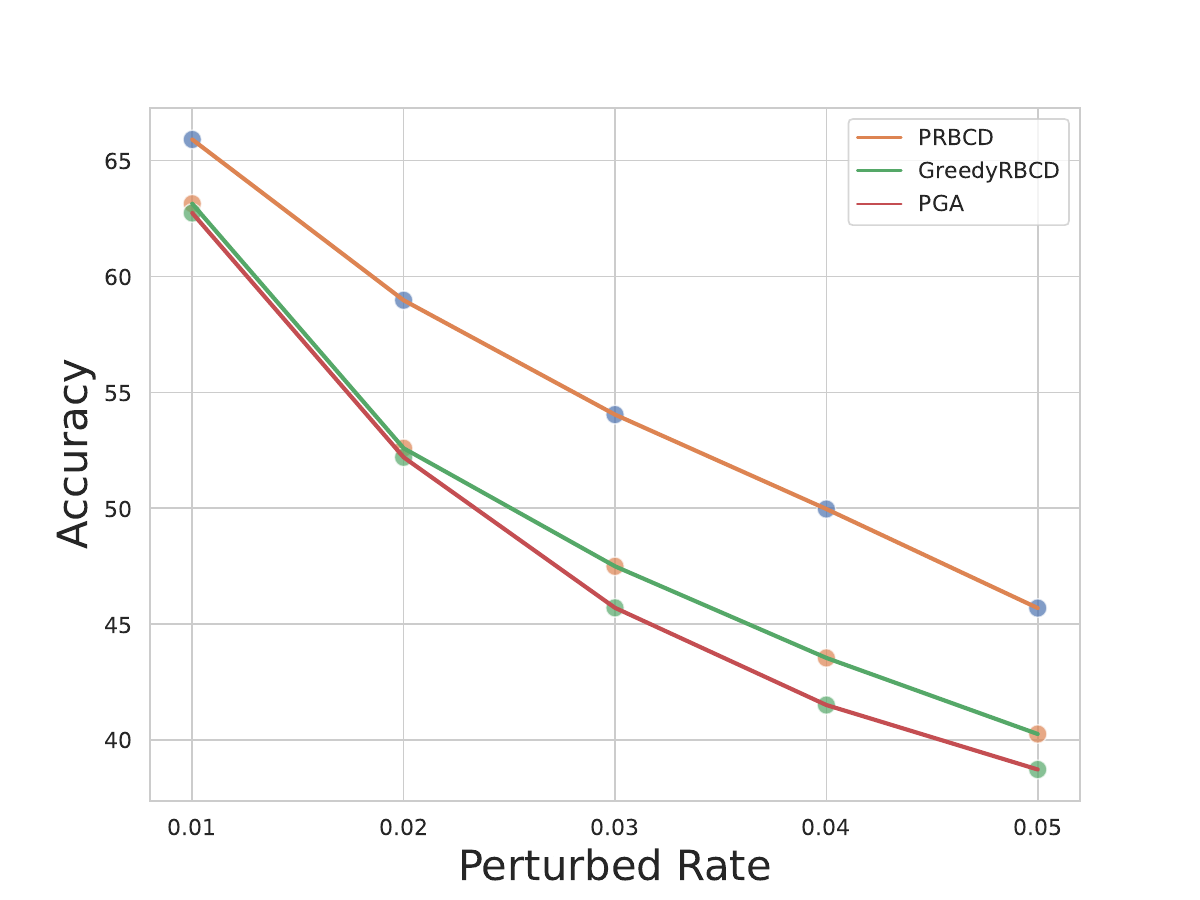}
  %\caption{fig2}
  \end{minipage}
  }%
  \centering
  \vspace{2ex}
  \caption{Global attack effects
    under different attack budgets (i.e., from 1\% to 5\% perturbed edges).}
  \label{experiments: range attack}
  \vspace{-3ex}
\end{figure}

% \subsubsection{1\%-5\% Range Attacks with Different Victim Models}
We vary the attack budget from 1\%-5\% and evaluate the attack effect at different perturbation levels, as depicted in Figure~\ref{experiments: range attack}.
We select PGD, Greedy, PRBCD, and GreedyRBCD as the baseline methods, considering their attack effects.

The results demonstrate the effectiveness of \mymodel{} across various attack budgets, even with very small attack budgets. 
For densely connected datasets (e.g., Pubmed) with larger attack budgets, the improvement of attack effect of \mymodel{} tends to exhibit a slower rate. 
Moreover, it is worth noting that attacks with smaller budgets are more realistic.
Excessive creation of fake edges may disrupt the properties of the original graph, such as the node degree distribution \cite{Nettack, Mettack}.
%

% Overall, the findings illustrate that our proposed \mymodel{} demonstrates its effectiveness across various attack budgets.

\subsection{Comparison of Attack Efficiency (RQ3)}
\begin{figure}[t]
  \setlength{\abovecaptionskip}{0.2cm}
  \centering
  \subfigure[Cora]{
  \hspace{-4mm}
  \begin{minipage}[t]{0.5\linewidth}
  \centering
  \includegraphics[width=1.85in]{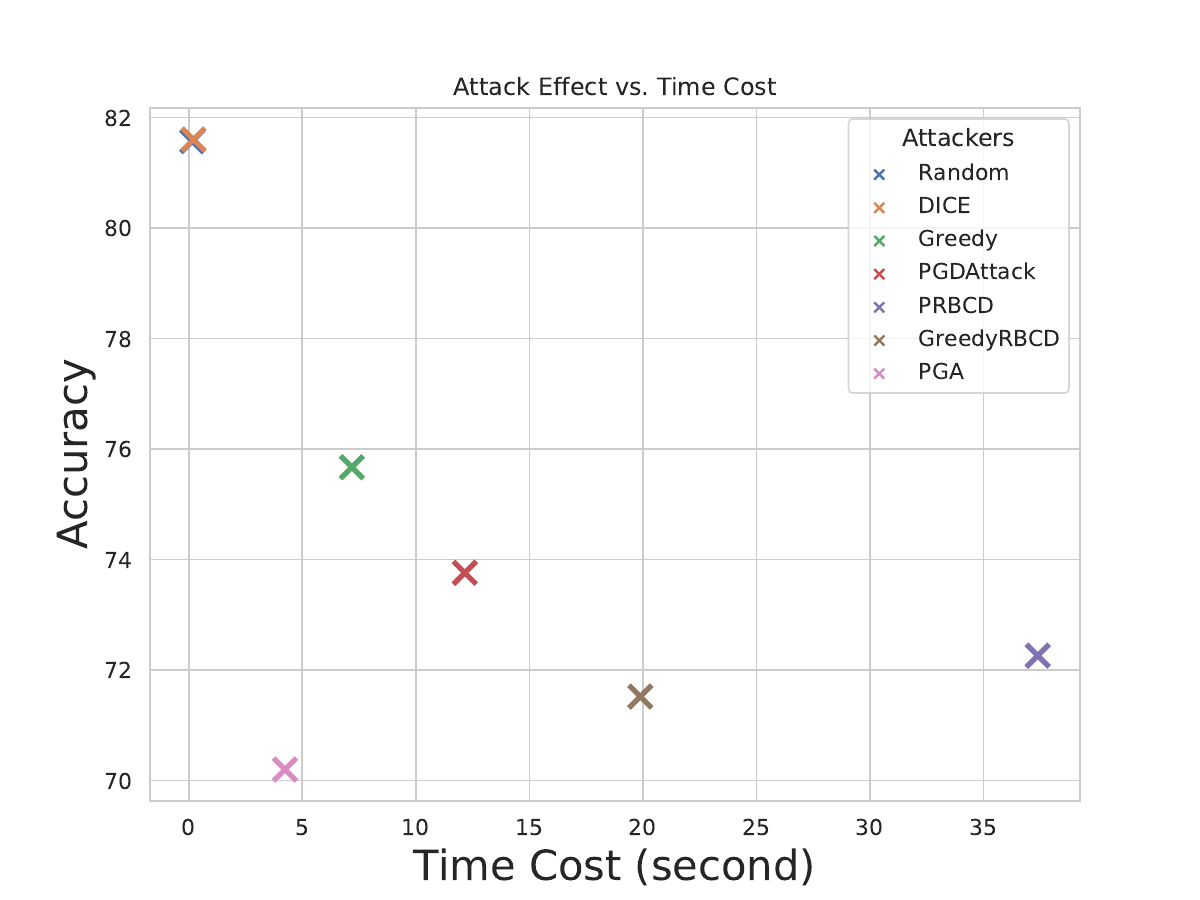}
  %\caption{fig2}
  \end{minipage}
  }%
  \subfigure[Citeseer]{
  \begin{minipage}[t]{0.5\linewidth}
  \centering
  \includegraphics[width=1.85in]{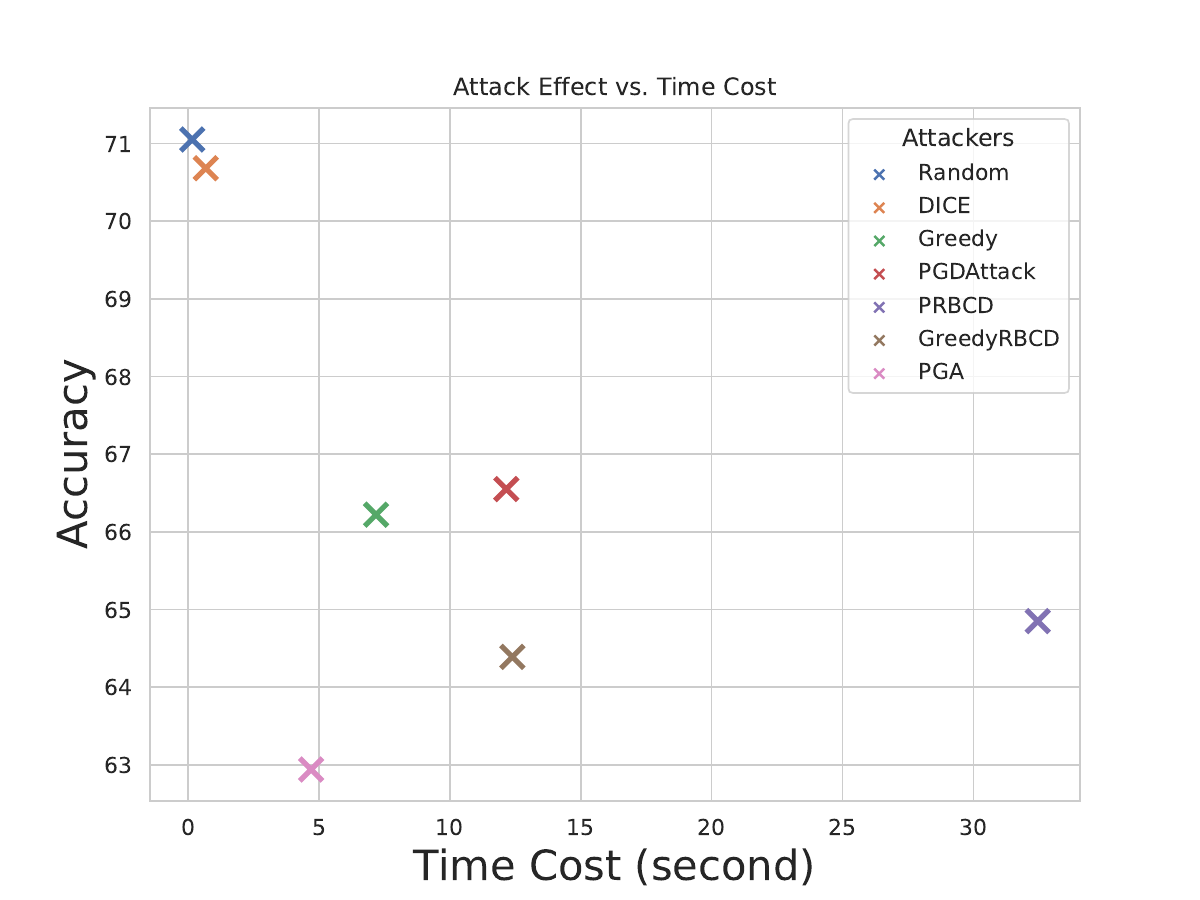}
  %\caption{fig2}
  \end{minipage}
  }%
  \quad
  \centering
  \vspace{-4mm}
  \subfigure[CoraML]{
  \hspace{-4mm}
  \begin{minipage}[t]{0.5\linewidth}
  \centering
  \includegraphics[width=1.85in]{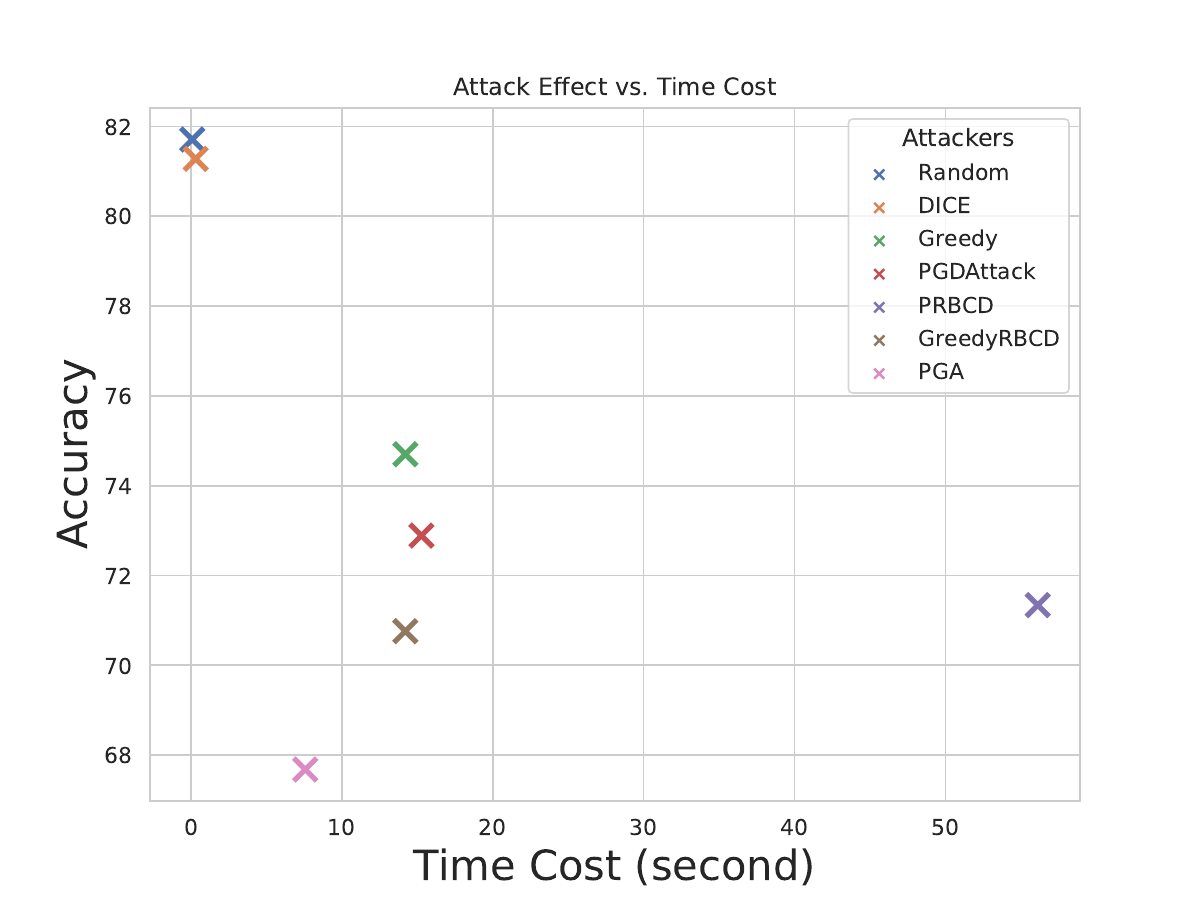}
  %\caption{fig2}
  \end{minipage}
  }%
  \subfigure[Pubmed]{
  \begin{minipage}[t]{0.5\linewidth}
  \centering
  \includegraphics[width=1.85in]{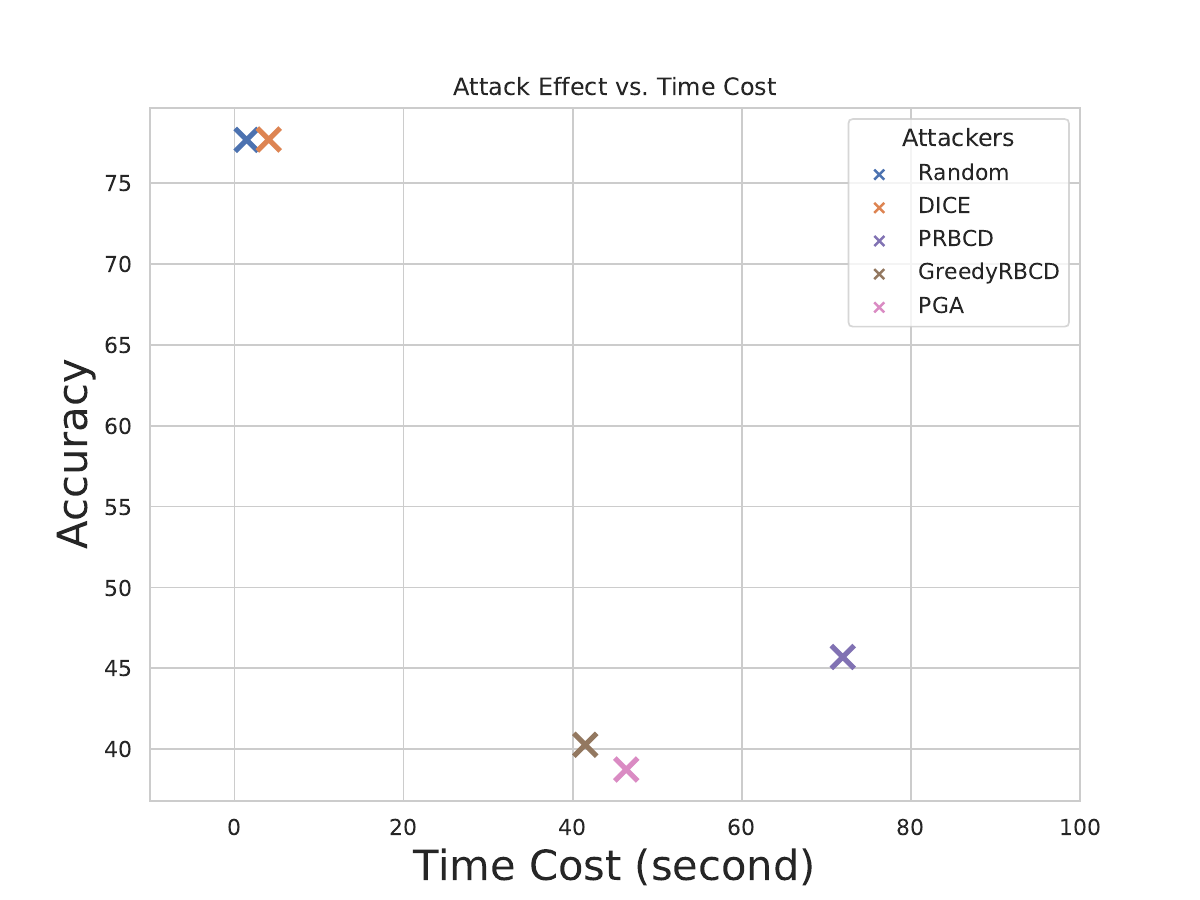}
  %\caption{fig2}
  %\label{figs-effiency-pubmed}
  \end{minipage}
  }%
  \centering
  \vspace{2ex}
  \caption{Comparison of the attack effects between different attack methods. 
  %The above four figures are the experimental results of the four data sets. 
  The x-axis records the time spent by the attack method (in seconds), and the y-axis is the classification accuracy of the victim model after being attacked. 
  The closer to the lower left corner, the better. 
  }
  \label{experiments: time cost}
  \vspace{-2ex}
\end{figure}

Furthermore, we evaluate the attack efficiency in terms of the running time consumed by different attack methods. The results in Figure \ref{experiments: time cost} show that the proposed \mymodel{} exhibits lower time costs compared to existing attack methods in most cases, indicating better attack efficiency, particularly on larger datasets. Greedy-based methods demonstrate higher efficiency when the dataset is small, but their advantages diminish as the dataset size increases. Although \mymodel{} is also based on a greedy strategy, it does not require gradient calculation for every candidate edge, as it focuses only on a subset of attack targets. This characteristic enables efficient adversarial attacks on larger datasets. 

In addition, although Random and DICE have a low time overhead, the attack effect is not ideal. PGD and Greedy experience a significant increase in time consumption on Pubmed, as it needs to consider every potential edge. Specifically, PGD's attack on Pubmed takes more than 30 minutes, and Greedy takes 4 hours. Since the attack time overhead of PGD and Greedy on Pubmed far exceeds other methods, we do not show PGDAttack and Greedy in~Figure~\ref{experiments: time cost}(d). In contrast, PRBCD and its variant method, GreedyRBCD, employ random block gradient descent, leading to a significant reduction in attack time. Further improvements in efficiency and performance could potentially be achieved by combining \mymodel{} with the random block gradient descent. Exploring this possibility is left for future work.

\subsection{Ablation Study (RQ4)}
We conduct ablation studies to study the 
  effectiveness of the attack target selection policy. 
  We compare \mymodel{} with its variants: 
  (1) \textbf{Random}: randomly select attack targets.
  (2) \textbf{Without Degree}: use only margin filter. 
  (3) \textbf{Without Margin}: use only degree filter. 
  (4) \textbf{Without PreF}: remove the preprocessing filter and directly select attack targets. 

The comparison results are shown in Table \ref{Ablation1}. The attack effect of \mymodel{} is the best when compared with all other variants in scenarios with different datasets and different victim models, which indicates that each component in the hierarchical target selection policy is necessary.

\begin{table}[t]
  \caption{Comparison of attack effects between \mymodel{} and its variants.
  %for attack target selection policy.
  }
  \resizebox{0.47\textwidth}{!}{
          \begin{tabular}{c|c|ccccc}
                \toprule \toprule
        Dataset 
        & Selection Policy                   
            & GCN                       
            & GAT                     
            & SGC                         
            & GraphSAGE
            & APPNP
            % & RGCN                       
            % & MedianGCN                     
            % & Jaccard                         
            % & Grand
            % & GNNGuard
            % & Avg.
            \\
            \midrule
            \multirow{5}{*}{Cora} 
                & Random Select
                    & 69.98±0.55  % GCN
                    & 75.30±0.51  % GAT
                    & 70.56±0.48  % SGC
                    & 75.20±0.78  % SAGE
                    & 77.62±0.51  % APPNP
                    
                    % & 71.92±1.24  % RGCN
                    % & 71.88±0.41  % Median
                    % & 67.56±0.82  % Jaccard
                    % & 70.54±0.34  % Grand
                    % & 75.26±0.39  % GNNGuard
                    \\
                & Only Margin 
                    & 67.62±1.05  % GCN
                    & 73.46±0.27  % GAT
                    & 68.56±0.59  % SGC
                    & 72.94±0.77  % SAGE
                    & 75.12±0.85  % APPNP
                    
                    % & 71.92±1.24  % RGCN
                    % & 71.88±0.41  % Median
                    % & 67.56±0.82  % Jaccard
                    % & 70.54±0.34  % Grand
                    % & 75.26±0.39  % GNNGuard
                    \\
                & Only Degree 
                    & 68.26±0.48  % GCN
                    & 72.24±0.74  % GAT
                    & 68.34±0.84  % SGC
                    & 72.18±0.70  % SAGE
                    & 75.46±0.68  % APPNP
                    
                    % & 71.92±1.24  % RGCN
                    % & 71.88±0.41  % Median
                    % & 67.56±0.82  % Jaccard
                    % & 70.54±0.34  % Grand
                    % & 75.26±0.39  % GNNGuard
                    \\
                & Without PreF 
                    & 68.84±0.82  % GCN
                    & 72.58±0.68  % GAT
                    & 69.02±0.56  % SGC
                    & 72.56±0.83  % SAGE
                    & 74.70±1.02  % APPNP
                    
                    % & 71.92±1.24  % RGCN
                    % & 71.88±0.41  % Median
                    % & 67.56±0.82  % Jaccard
                    % & 70.54±0.34  % Grand
                    % & 74.70±1.02  % GNNGuard
                    \\
                & \textbf{PGA} 
                    & \textbf{66.26±0.59}  % GCN
                    & \textbf{71.22±0.42}  % GAT
                    & \textbf{67.50±0.21}  % SGC
                    & \textbf{71.42±0.67}  % SAGE
                    & \textbf{74.60±0.64}  % APPNP
                    
                    % & 71.92±1.24  % RGCN
                    % & 71.88±0.41  % Median
                    % & 67.56±0.82  % Jaccard
                    % & 70.54±0.34  % Grand
                    % & 75.26±0.39  % GNNGuard
                    \\

                    \midrule
            \multirow{5}{*}{Citeseer} 
                & Random Select 
                    & 63.36±1.04  % GCN
                    & 65.18±0.61  % GAT
                    & 64.18±0.55 % SGC
                    & 67.24±0.87  % SAGE
                    & 65.98±1.32  % APPNP
                    
                    % & 71.92±1.24  % RGCN
                    % & 71.88±0.41  % Median
                    % & 67.56±0.82  % Jaccard
                    % & 70.54±0.34  % Grand
                    % & 75.26±0.39  % GNNGuard
                    \\
                & Only Margin 
                    & 63.42±0.20  % GCN
                    & 65.78±0.69  % GAT
                    & 65.88±0.38  % SGC
                    & 66.22±0.53  % SAGE
                    & 66.78±0.37  % APPNP
                    
                    % & 71.92±1.24  % RGCN
                    % & 71.88±0.41  % Median
                    % & 67.56±0.82  % Jaccard
                    % & 70.54±0.34  % Grand
                    % & 75.26±0.39  % GNNGuard
                    \\
                & Only Degree 
                    & 61.60±0.58  % GCN
                    & 63.20±1.16  % GAT
                    & 62.76±0.59  % SGC
                    & 65.08±0.60  % SAGE
                    & 64.34±0.45  % APPNP
                    
                    % & 71.92±1.24  % RGCN
                    % & 71.88±0.41  % Median
                    % & 67.56±0.82  % Jaccard
                    % & 70.54±0.34  % Grand
                    % & 75.26±0.39  % GNNGuard
                    \\
                & Without PreF 
                    & 62.28±1.13  % GCN
                    & 63.70±0.79  % GAT
                    & 63.56±0.99  % SGC
                    & 65.44±1.25  % SAGE
                    & 64.42±0.71  % APPNP
                    
                    % & 71.92±1.24  % RGCN
                    % & 71.88±0.41  % Median
                    % & 67.56±0.82  % Jaccard
                    % & 70.54±0.34  % Grand
                    % & 74.70±1.02  % GNNGuard
                    \\
                & \textbf{PGA} 
                    & \textbf{61.12±0.97}  % GCN
                    & \textbf{62.86±0.65}  % GAT
                    & \textbf{62.46±1.15}  % SGC
                    & \textbf{64.12±0.60}  % SAGE
                    & \textbf{64.14±0.96}  % APPNP
                    
                    % & 71.92±1.24  % RGCN
                    % & 71.88±0.41  % Median
                    % & 67.56±0.82  % Jaccard
                    % & 70.54±0.34  % Grand
                    % & 75.26±0.39  % GNNGuard
                    \\

                    \midrule
            \multirow{5}{*}{CoraML} 
                & Random Select 
                    & 67.44±0.80  % GCN
                    & 73.58±0.47  % GAT
                    & 67.76±0.97  % SGC
                    & 73.34±0.81  % SAGE
                    & 75.04±0.81  % APPNP
                    
                    % & 71.92±1.24  % RGCN
                    % & 71.88±0.41  % Median
                    % & 67.56±0.82  % Jaccard
                    % & 70.54±0.34  % Grand
                    % & 75.26±0.39  % GNNGuard
                    \\
                & Only Margin 
                    & 63.94±0.67  % GCN
                    & 71.32±1.11  % GAT
                    & 65.00±0.64  % SGC
                    & 70.18±0.74  % SAGE
                    & 72.42±0.87  % APPNP
                    
                    % & 71.92±1.24  % RGCN
                    % & 71.88±0.41  % Median
                    % & 67.56±0.82  % Jaccard
                    % & 70.54±0.34  % Grand
                    % & 75.26±0.39  % GNNGuard
                    \\
                & Only Degree 
                    & 63.80±0.54  % GCN
                    & 71.00±1.66  % GAT
                    & 65.42±0.41  % SGC
                    & 70.84±0.45  % SAGE
                    & 73.12±1.07  % APPNP
                    
                    % & 71.92±1.24  % RGCN
                    % & 71.88±0.41  % Median
                    % & 67.56±0.82  % Jaccard
                    % & 70.54±0.34  % Grand
                    % & 75.26±0.39  % GNNGuard
                    \\
                & Without PreF 
                    & 64.16±0.52  % GCN
                    & 70.76±1.77  % GAT
                    & 65.88±0.84  % SGC
                    & 69.90±0.61  % SAGE
                    & 72.70±1.33  % APPNP
                    
                    % & 71.92±1.24  % RGCN
                    % & 71.88±0.41  % Median
                    % & 67.56±0.82  % Jaccard
                    % & 70.54±0.34  % Grand
                    % & 74.70±1.02  % GNNGuard
                    \\
                & \textbf{PGA} 
                    & \textbf{62.62±0.29}  % GCN
                    & \textbf{70.16±1.08}  % GAT
                    & \textbf{64.24±0.62}  % SGC
                    & \textbf{69.42±0.66}  % SAGE
                    & \textbf{71.96±0.71}  % APPNP
                    
                    % & 71.92±1.24  % RGCN
                    % & 71.88±0.41  % Median
                    % & 67.56±0.82  % Jaccard
                    % & 70.54±0.34  % Grand
                    % & 75.26±0.39  % GNNGuard
                    \\

 \midrule
            \multirow{4}{*}{Pubmed} 
                & Random Select 
                    & 44.56±1.34  % GCN
                    & 52.48±1.24  % GAT
                    & 48.66±1.32  % SGC
                    & 58.10±0.99  % SAGE
                    & 54.34±1.46  % APPNP
                    
                    % & 71.92±1.24  % RGCN
                    % & 71.88±0.41  % Median
                    % & 67.56±0.82  % Jaccard
                    % & 70.54±0.34  % Grand
                    % & 75.26±0.39  % GNNGuard
                    \\
                & Only Margin 
                    & 33.02±0.81  % GCN
                    & 40.22±1.16  % GAT
                    & 38.00±0.40  % SGC
                    & 48.58±2.11  % SAGE
                    & 44.78±2.38  % APPNP
                    
                    % & 71.92±1.24  % RGCN
                    % & 71.88±0.41  % Median
                    % & 67.56±0.82  % Jaccard
                    % & 70.54±0.34  % Grand
                    % & 75.26±0.39  % GNNGuard
                    \\
                & Only Degree 
                    & 30.66±0.55  % GCN
                    & 38.02±2.18  % GAT
                    & 35.66±1.57  % SGC
                    & 47.80±1.86  % SAGE
                    & 43.60±3.00  % APPNP
                    
                    % & 71.92±1.24  % RGCN
                    % & 71.88±0.41  % Median
                    % & 67.56±0.82  % Jaccard
                    % & 70.54±0.34  % Grand
                    % & 75.26±0.39  % GNNGuard
                    \\
                & Without PreF 
                    & 31.02±0.36  % GCN
                    & 38.90±1.58  % GAT
                    & 35.98±1.31  % SGC
                    & 47.88±2.32  % SAGE
                    & 43.38±2.76  % APPNP
                    
                    % & 71.92±1.24  % RGCN
                    % & 71.88±0.41  % Median
                    % & 67.56±0.82  % Jaccard
                    % & 70.54±0.34  % Grand
                    % & 74.70±1.02  % GNNGuard
                    \\
                & \textbf{PGA} 
                    & \textbf{29.98±0.40}  % GCN
                    & \textbf{37.36±1.40}  % GAT
                    & \textbf{35.58±1.38}  % SGC
                    & \textbf{47.68±2.21}  % SAGE
                    & \textbf{43.12±2.84}  % APPNP
                    
                    % & 71.92±1.24  % RGCN
                    % & 71.88±0.41  % Median
                    % & 67.56±0.82  % Jaccard
                    % & 70.54±0.34  % Grand
                    % & 75.26±0.39  % GNNGuard
                    \\
                
                \bottomrule
                \bottomrule

          \end{tabular}
  }
\label{Ablation1}%
\vspace{-3ex}
\end{table}%

In \mymodel{}, we primarily focus on pre-selecting nodes deemed vulnerable to attack. 
It is also essential to investigate the consequences of attack nodes considered robust or less vulnerable. As depicted in Table \ref{Ablation2}, the attack effects between the two target selection policies (i.e., attacking weak nodes or robust nodes) exhibit significant differences. 
Attacking weak nodes shows much better attack effect.
This observation highlights the crucial role of selecting the appropriate attack targets, 
%
%It emphasizes that the choice of attack targets 
which has a substantial impact on the overall attack effect. 

\begin{table}[t]
  \caption{Comparison of attacking weak nodes and robust nodes.}
  \resizebox{0.47\textwidth}{!}{
          \begin{tabular}{c|ccccc}
                \toprule \toprule
                           
            & GCN                       
            & GAT                     
            & SGC                         
            & GraphSAGE
            & APPNP
            \\
            \midrule
                % Att Weak
                %     & 66.26±0.59  % GCN
                %     & 71.22±0.42  % GAT
                %     & 67.50±0.21  % SGC
                %     & 67.50±0.21  % SAGE
                %     & 74.60±0.64  % APPNP
                %     \\
                % Att Robust 
                %     & 75.48±0.79  % GCN
                %     & 80.22±0.81  % GAT
                %     & 75.28±0.74  % SGC
                %     & 79.92±0.12  % SAGE
                %     & 81.34±0.89  % APPNP
                %     \\
                % Clean Acc 
                %     & 82.02±0.52  % GCN
                %     & 83.12±0.49  % GAT
                %     & 80.74±0.10  % SGC
                %     & 81.52±0.33  % SAGE
                %     & 83.92±0.31  % APPNP
                    
                %     \\
                 Clean Acc 
                    & 82.02±0.52  % GCN
                    & 83.12±0.49  % GAT
                    & 80.74±0.10  % SGC
                    & 81.52±0.33  % SAGE
                    & 83.92±0.31  % APPNP
                    \\
                Attack Robust 
                    & 75.48±0.79  % GCN
                    & 80.22±0.81  % GAT
                    & 75.28±0.74  % SGC
                    & 79.92±0.12  % SAGE
                    & 81.34±0.89  % APPNP
                    \\
                Attack Weak
                    & \textbf{66.26±0.59}  % GCN
                    & \textbf{71.22±0.42}  % GAT
                    & \textbf{67.50±0.21}  % SGC
                    & \textbf{67.50±0.21}  % SAGE
                    & \textbf{74.60±0.64}  % APPNP
                    \\
                \bottomrule
                \bottomrule

          \end{tabular}
  }
\label{Ablation2}%
\vspace{-3ex}
\end{table}%

\subsection{Unnoticeable Perturbations (RQ5)} 

\begin{figure}[t]
    \setlength{\abovecaptionskip}{-0.1cm}
    \centering
    \subfigure[Cora]{
    \begin{minipage}[t]{0.48\linewidth}
    \centering
    \includegraphics[width=1.65in]{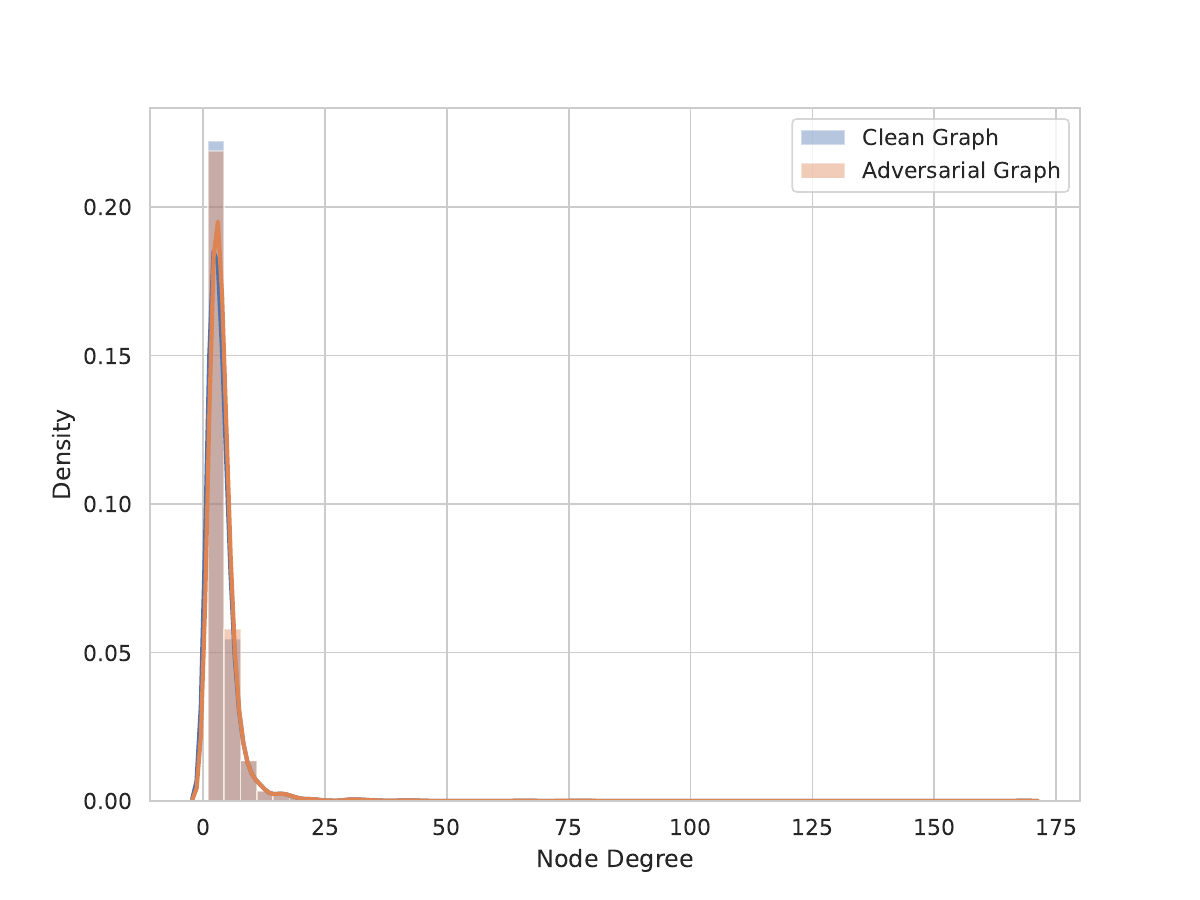}
    \end{minipage}
    }%
    \subfigure[Citeseer]{
    \begin{minipage}[t]{0.48\linewidth}
    \centering
    \includegraphics[width=1.65in]{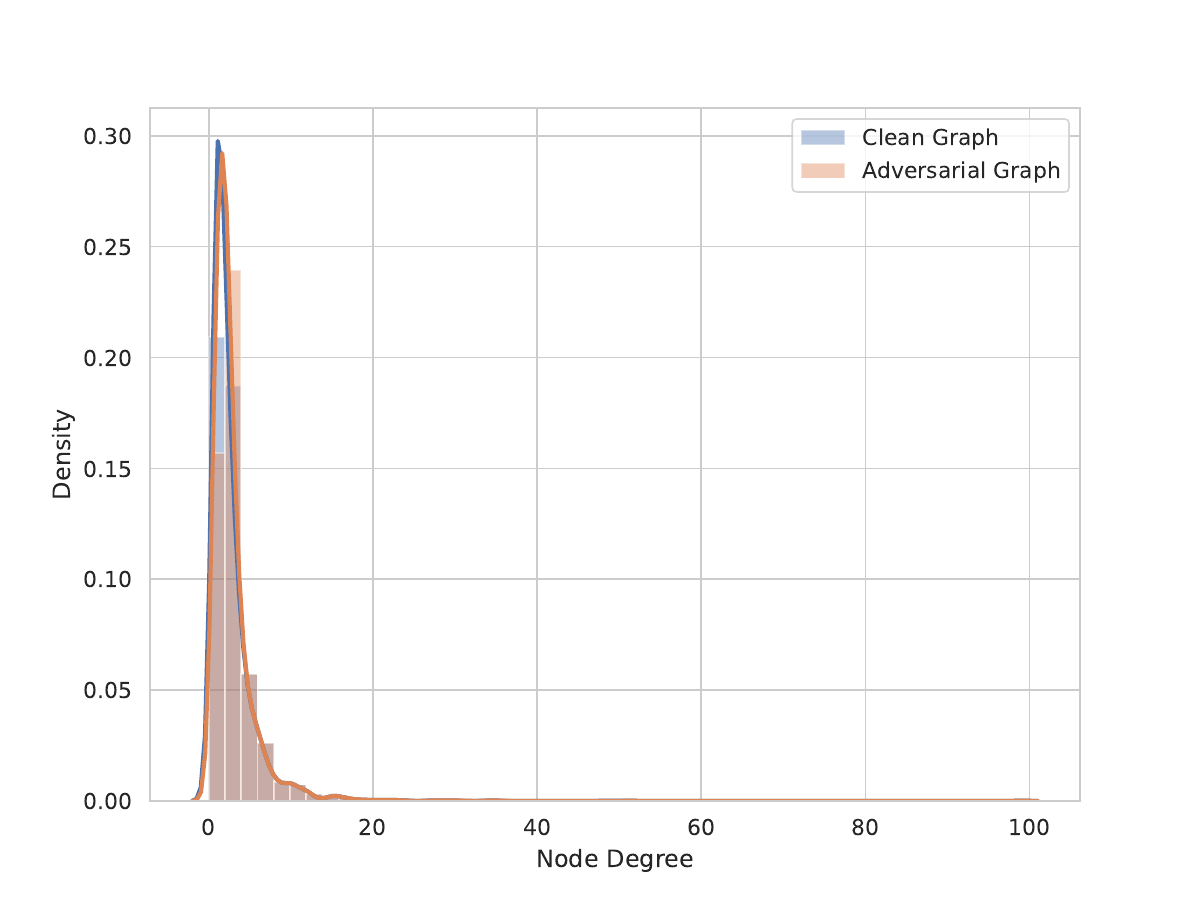}
    \end{minipage}
    }%
    \quad
    \centering
    \subfigure[CoraML]{
    \begin{minipage}[t]{0.48\linewidth}
    \centering
    \includegraphics[width=1.65in]{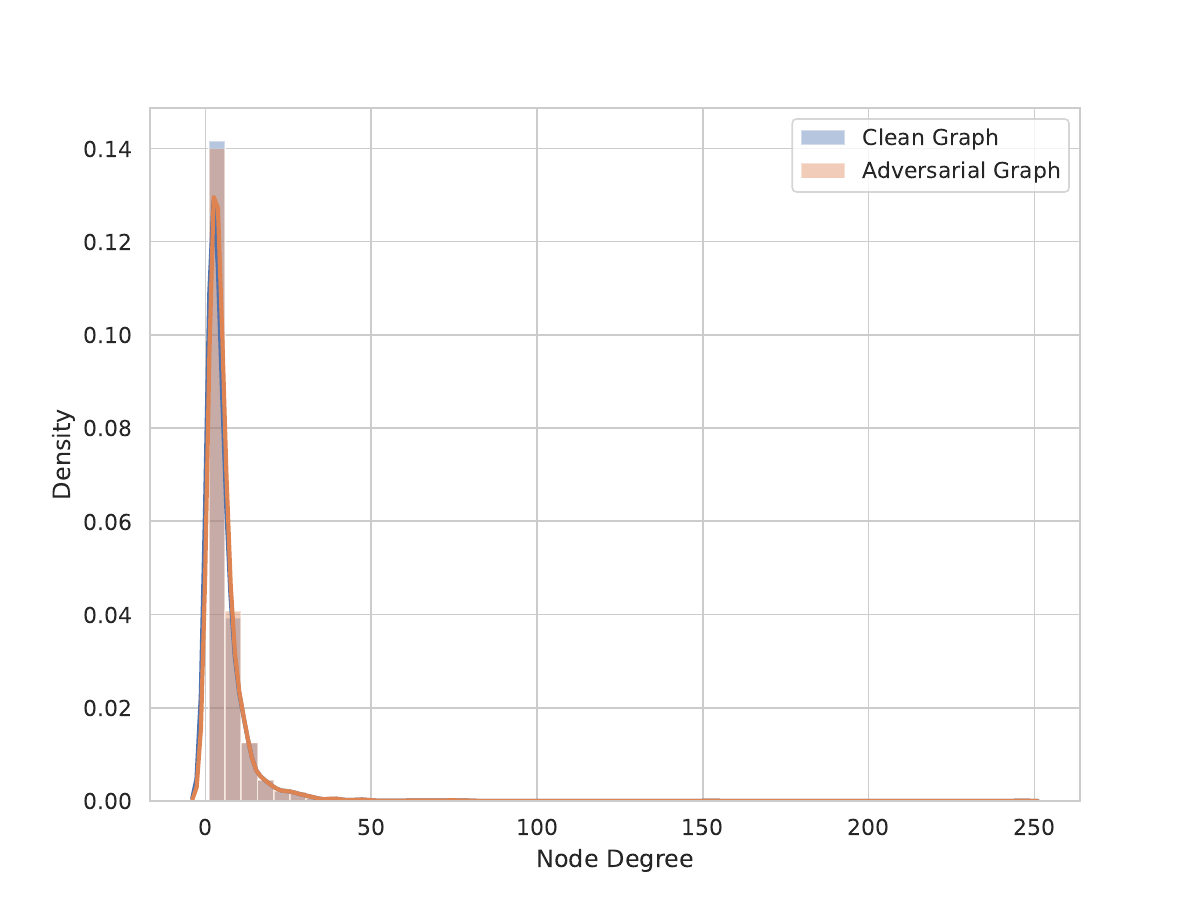}
    \end{minipage}
    }%
    \subfigure[Pubmed]{
    \begin{minipage}[t]{0.48\linewidth}
    \centering
    \includegraphics[width=1.65in]{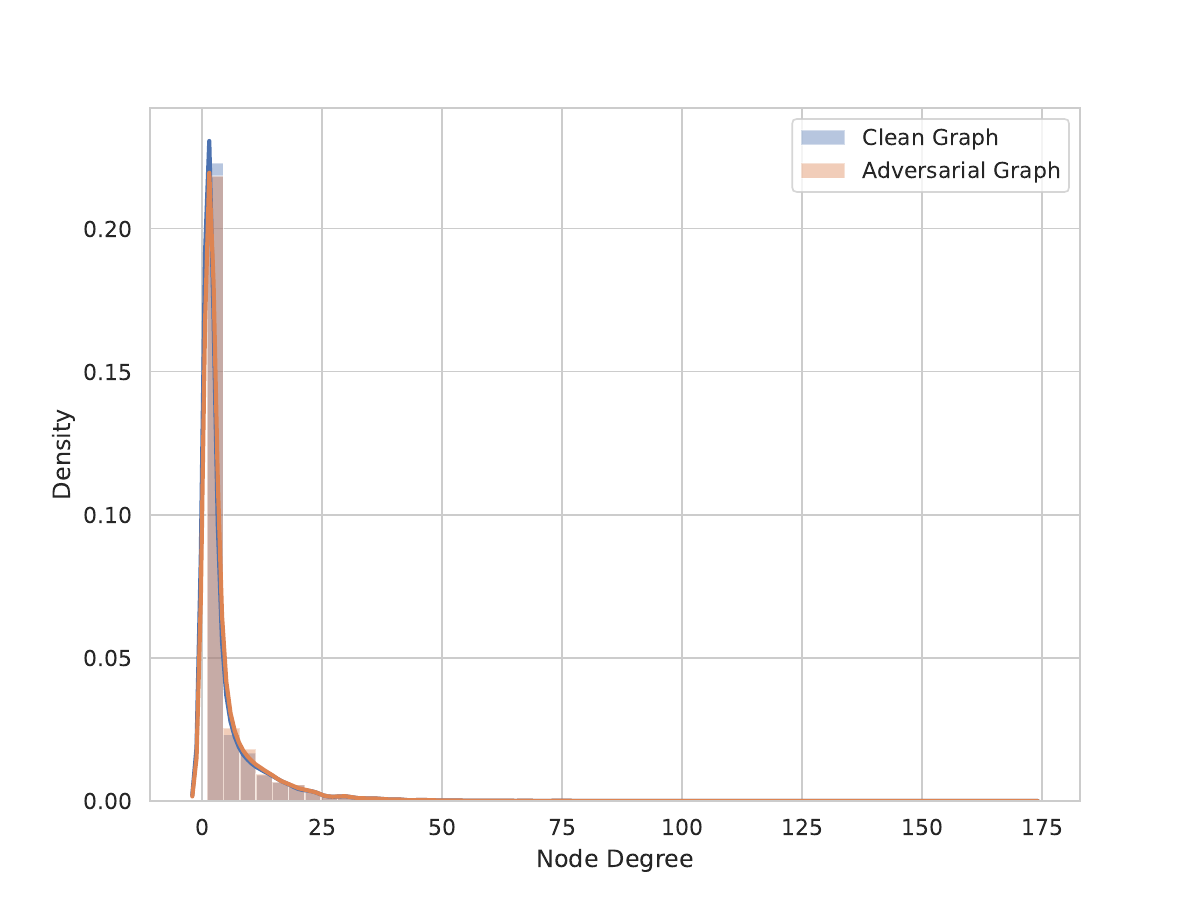}
    \end{minipage}
    }%
    \centering
    \vspace{1ex}
    \caption{Comparison of node degree distribution between adversarial graphs and original graphs. }
    \label{experiments: degree distribution}
    \vspace{-3ex}
  \end{figure}

In this section, we visualize the graphs before and after the attack, as shown in Figure \ref{experiments: degree distribution}. 
Unlike adversarial attacks in the field of computer vision, distinguishing the differences between adversarial and original graphs is challenging in the graph domain.
To some extent, the node degree distribution can measure the graph's changes \cite{Nettack}. 
If the node degree distribution of the attacked graph does not significantly deviate from that of the original graph, it indicates that the attack does not cause catastrophic disruption to the properties of the original graph \cite{Nettack, Mettack}. 
% In that case, it indicates that the attack does not cause catastrophic disruption to the properties of the original graph \cite{Nettack, Mettack}.

As depicted in Figure \ref{experiments: degree distribution}, the proposed \mymodel{} generates perturbations on the original graph that are visually indistinguishable. 
The node degree distribution of the attacked graph remains similar to that of the original graph, indicating that the attack does not inflict severe damage to the graph's properties. 
These results highlight the stealthiness of \mymodel{}.

\subsection{Effectiveness Under the Poisoning Attack Setting (RQ6)}
Furthermore, we carry out a series of poisoning attack experiments. Currently, we have not implemented any specific design for the poisoning attack. Instead, we employ the adversarial graph obtained by the evasion attack  and apply it to the poisoning attack scenario. For the purpose of comparison, we select five methods with similar attack effects: Random, DICE, Greedy, PRBCD, GreedyRBCD, and \mymodel{}. 
We select 10 different GNN models as victims. 
The results in Table~\ref{experiments: poisoning attack} shows that \mymodel{} can also achieve better attack effects under the poisoning attack setting, demonstrating the successful migration of our method to the poisoning attack scenario.
%
% %
% However, our method does not work well in attacking GNNGuard on Pubmed.
% %
% This is because that the GNN model trained on Pubmed is generally not robust enough and close to 60\% of nodes (1000 test nodes) cannot defend against only one false neighbor in the targeted attack. \mymodel{} is likely to lose some vulnerable node choices. 
% %
% In addition, since Pubmed is a 3-class task, the second and third elements of the output logit may have considerable interference effects on the final classification result.

\begin{table*}[t]
  \centering
  \caption{Poisoning attack effect on different models (the lower, the better) under 5\% perturbed edges. Bold and underline indicate the best and the second best, respectively. }
  \resizebox{\textwidth}{!}{
         \begin{tabular}{c|c|ccccc|ccccc}
                \toprule
                \textbf{Dataset}
                & \textbf{Attackers}     
                & \textbf{GCN}  
                & \textbf{GAT} 
                & \textbf{SGC}  
                & \textbf{GraphSAGE} 
                & \textbf{APPNP} 
                & \textbf{RGCN}  
                & \textbf{MedianGCN} 
                & \textbf{Jaccard}  
                & \textbf{Grand} 
                & \textbf{GNNGuard} \\
                \midrule
                \multirow{5}{*}{\textbf{Cora}} 
                    
                    & \textbf{Greedy}   
                    &74.78±1.12
&75.86±0.90
&73.16±1.16
&75.28±0.77
&77.20±0.94
&74.94±0.87
&75.26±0.63
&75.20±0.85
&78.50±0.86
&76.34±0.71\\

                    & \textbf{tanh-PGD}   
                    &71.38±1.11
&73.68±1.33
&71.00±0.61
&73.36±0.52
&75.32±1.08
&72.06±0.68
&75.16±0.89
&74.40±0.79
&76.66±0.64
&76.32±0.58\\

                    & \textbf{PRBCD}   
                    &70.36±0.60
&71.98±0.59
&70.22±0.72
&\underline{71.86±0.79}
&\underline{74.70±0.55}
&71.24±0.80
&75.34±0.95
&75.14±0.88
&\underline{75.98±0.74}
&76.36±0.72\\

                    & \textbf{GreedyRBCD}   
                    &\underline{68.60±0.61}
&\underline{71.26±0.82}
&\underline{68.14±0.85}
&71.90±0.45
&75.36±1.10
&\underline{70.06±0.65}
&\textbf{74.76±0.83}
&\underline{74.04±0.62}
&76.86±0.58
&\textbf{75.64±0.62} \\

                    & \textbf{PGA}   
                    & \textbf{65.90±0.72}
& \textbf{70.44±0.66}
& \textbf{66.42±0.60}
& \textbf{70.40±0.97}
&\textbf{73.68±0.54}
&\textbf{67.58±1.34}
&\underline{74.88±0.64}
&\textbf{72.56±0.95}
&\textbf{75.14±0.80}
&\underline{75.94±0.63} \\

            \midrule
                \multirow{5}{*}{\textbf{Citeseer}} 
                    
                    & \textbf{Greedy}   
                    &66.00±0.79
&66.18±0.53
&66.80±0.28
&68.16±0.60
&66.72±0.71
&65.98±0.49
&67.56±0.96
&68.94±0.69
&69.80±0.54
&69.58±0.37\\

                    & \textbf{tanh-PGD}   
                    &65.06±0.85
&65.26±1.11
&65.78±0.82
&66.88±0.52
&65.30±1.72
&64.36±0.68
&66.30±0.95
&67.74±1.60
&68.18±1.28
&68.14±0.29\\

                    & \textbf{PRBCD}   
                    &63.48±0.93
&\underline{63.98±1.12}
&64.32±0.53
&\underline{65.32±0.93}
&64.54±0.93
&63.68±0.53
&\underline{64.98±1.07}
&67.96±0.64
&\underline{66.92±0.37}
&\underline{67.40±0.40} \\

                    & \textbf{GreedyRBCD}   
                    &\underline{62.72±0.92}
&64.04±0.86
&\underline{64.28±1.60}
&65.38±1.17
&\underline{64.48±1.28}
&\underline{63.34±1.12}
&65.92±0.90
&\underline{67.58±0.79}
&69.38±0.86
&68.72±0.40\\

                    & \textbf{PGA}   
                   &\textbf{60.18±1.05}
&\textbf{61.58±0.99}
&\textbf{62.34±1.37}
&\textbf{63.54±0.86}
&\textbf{62.66±0.90}
&\textbf{60.88±0.76}
&\textbf{63.68±0.52}
&\textbf{67.48±1.12}
&\textbf{66.42±0.53}
&\textbf{67.14±0.49}\\                         

            \midrule
                \multirow{5}{*}{\textbf{CoraML}}

                    & \textbf{Greedy}   
                    &73.58±0.33
&74.10±0.63
&71.84±0.57
&73.98±0.45
&76.58±0.46
&73.56±0.66
&74.86±0.39
&75.96±0.79
&76.68±0.35
&76.80±0.67\\

                    & \textbf{tanh-PGD}   
                    &71.16±0.82
&72.82±0.78
&70.52±1.40
&72.18±1.21
&73.98±1.17
&71.16±0.98
&76.84±1.00
&75.10±1.26
&73.06±2.69
&77.28±1.14\\

                    & \textbf{PRBCD}   
                    &69.08±1.11
&71.18±1.43
&67.50±0.89
&\underline{70.92±0.96}
&\underline{72.76±1.55}
&68.58±1.03
&74.40±0.76
&\underline{73.88±0.57}
&\underline{72.34±1.23}
&\underline{75.96±0.97}\\

                    & \textbf{GreedyRBCD}   
                    &\underline{68.22±0.80}
&\underline{69.98±0.89}
&\underline{65.86±1.47}
&71.04±0.83
&75.58±1.09
&\underline{68.18±0.97}
&\underline{73.94±0.46}
&76.14±0.48
&75.62±1.30
&76.86±0.60\\

                    & \textbf{PGA}   
                    &\textbf{62.44±0.54}
&\textbf{68.16±0.90}
&\textbf{62.14±0.77}
&\textbf{67.86±0.72}
&\textbf{69.38±0.56}
&\textbf{62.66±0.22}
&\textbf{73.68±0.59}
&\textbf{72.30±0.57}
&\textbf{67.52±2.34}
&\textbf{75.48±0.69}\\

            \midrule
                \multirow{5}{*}{\textbf{Pubmed}} 
                    
                    & \textbf{Greedy}   
                    &58.34±1.09
&58.46±0.88
&57.44±0.75
&58.48±0.57
&59.56±0.77
&58.22±1.00
&57.74±0.76
&58.68±1.14
&59.48±0.64
&59.44±0.73\\

                    & \textbf{tanh-PGD}   
                    &68.98±1.29
&68.50±0.96
&67.92±1.34
&68.66±0.85
&70.60±1.35
&69.22±1.24
&69.72±1.00
&69.20±1.42
&70.86±1.53
&70.42±0.95\\

                    & \textbf{PRBCD}   
                    &40.64±1.38
&44.00±1.14
&43.42±1.79
&49.22±0.52
&46.80±1.75
&42.12±1.50
&47.48±1.11
&47.16±1.31
&50.18±1.38
&\textbf{53.48±2.62}\\

                    & \textbf{GreedyRBCD}   
                    &\underline{32.04±0.47}
&\underline{40.22±2.04}
&\underline{37.50±0.76}
&\underline{46.86±1.99}
&\underline{43.28±1.20}
&\underline{33.60±1.15}
&\textbf{45.16±1.68}
&\underline{43.18±0.45}
&\underline{44.44±1.37}
&\underline{58.86±2.28}\\

                    & \textbf{PGA}   
                    &\textbf{30.20±1.08}
&\textbf{38.30±2.60}
&\textbf{35.26±1.02}
&\textbf{46.62±2.24}
&\textbf{41.50±2.47}
&\textbf{31.94±1.08}
&\underline{45.24±0.95}
&\textbf{41.20±2.50}
&\textbf{44.30±2.04}
&61.10±2.32\\
                
                    \bottomrule
         \end{tabular}
  }
\label{experiments: poisoning attack}%
\end{table*}%

\subsection{Hyperparameter Analysis (RQ7)}
We further conduct two sets of hyperparameter experiments to investigate the impact of different filtering ratios and different surrogate models.
\subsubsection{Filtering ratio}
\begin{figure}[t]
  \setlength{\abovecaptionskip}{-0.1cm}
  \centering
  \subfigure[Cora]{
  \begin{minipage}[t]{0.48\linewidth}
  \centering
  \includegraphics[width=1.65in]{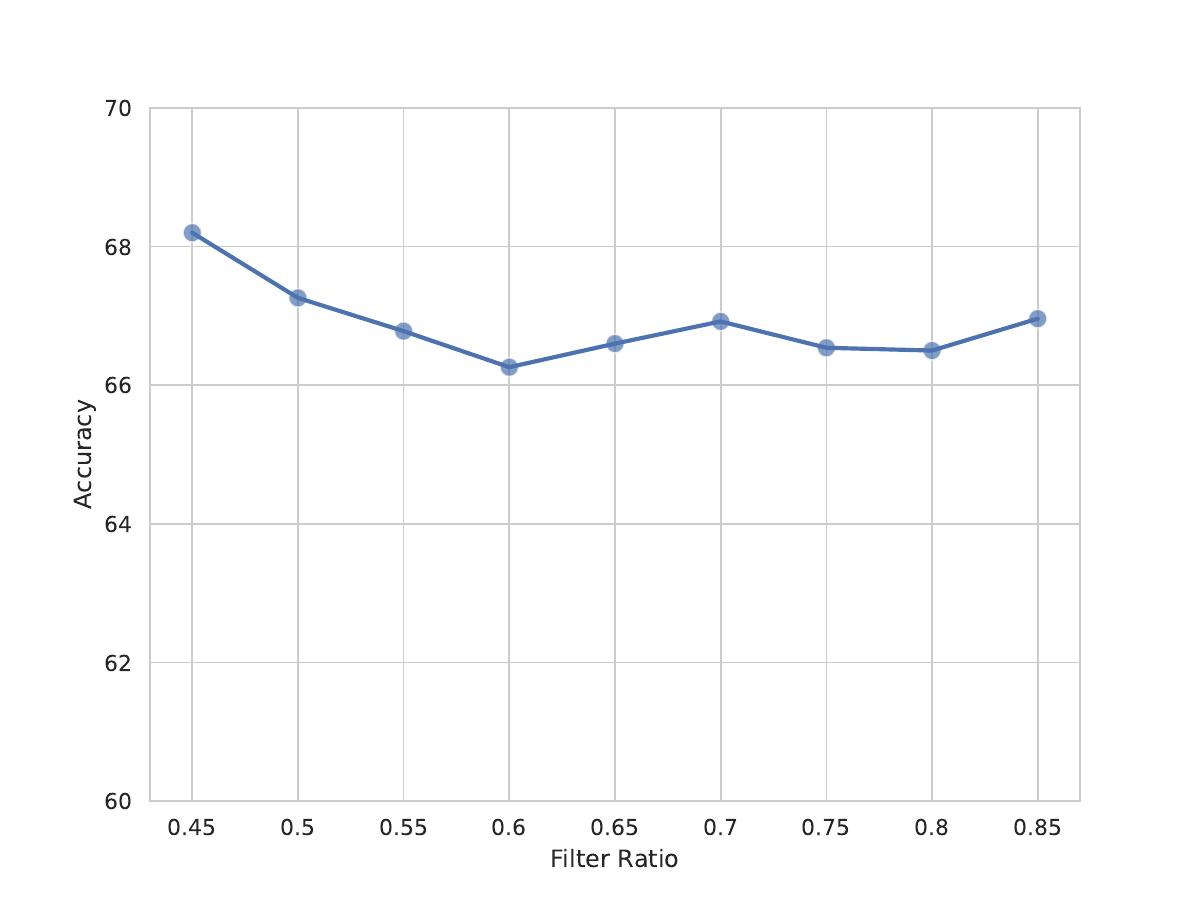}
  \end{minipage}
  }%
  \subfigure[Citeseer]{
  \begin{minipage}[t]{0.48\linewidth}
  \centering
  \includegraphics[width=1.65in]{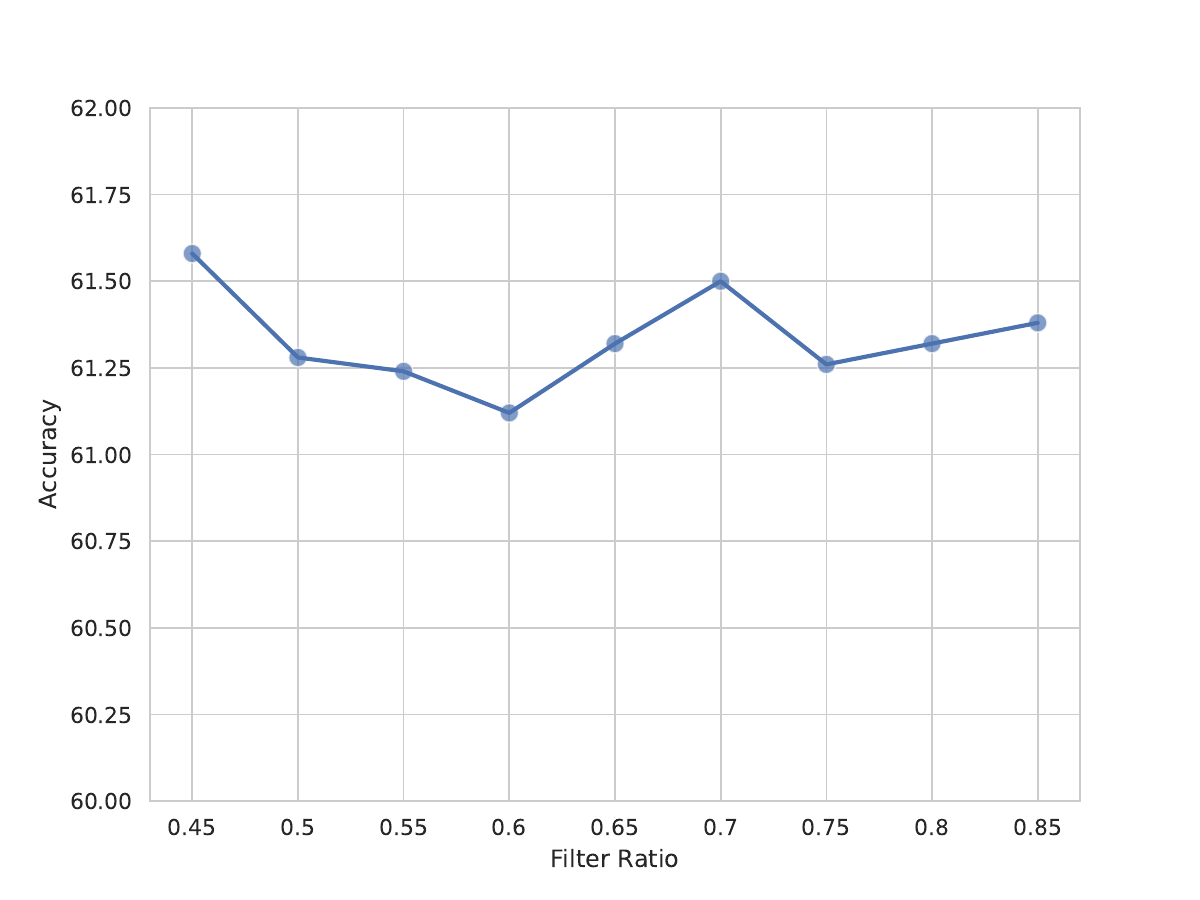}
  \end{minipage}
  }%
  \centering
  \vspace{1ex}
  \caption{The effect of the filter ratio on the Cora and Citeseer datasets.}
  \label{experiments: filter ratio hyperparameter}
  \vspace{-1ex}
\end{figure}

As mentioned earlier, we ultimately employ node degree and classification margin as indicators for identifying vulnerable nodes. To achieve this, we establish a filtering ratio, such as selecting the lowest ten percent based on degree and margin. To reduce the number of hyperparameters, we set the filter ratio for both node degree and margin to the same value. We vary the filtering ratio from 0.45 to 0.85. A larger filtering ratio indicates a higher number of potential attack targets. Ideally, when the filtering ratio is relatively small, the resulting attack targets would comprise the nodes that are more susceptible to adversarial attacks. From Figure \ref{experiments: filter ratio hyperparameter}, we observe that \mymodel{} is not highly sensitive to the chosen filtering ratio. Note that when using a smaller filtering ratio, the resulting attack targets will definitely be included in the attack targets obtained using a larger filtering ratio. Consequently, as the filtering ratio increases, the attack effect does not significantly deteriorate. 

Furthermore, we evaluate the robustness of the nodes selected by \mymodel{}. We previously performed targeted attack tests, and now we examine how resilient the nodes selected by \mymodel{} are when subjected to targeted attacks. Table \ref{hyper2} clearly demonstrates that \mymodel{} has effectively chosen more vulnerable nodes as attack objects, reinforcing this observation.
%

% The results in Table \ref{hyper2} reveal that the nodes chosen by \mymodel{} exhibit a lack of robustness, as they can be successfully attacked with a high success rate by simply adding a false neighbor. We counted the attack objects selected by \mymodel{} and determined the number of false neighbors required for the targeted attack.  

\begin{table}[t]
  \caption{Robustness of attack targets selected by \mymodel{}. 'Budget=1' means that only one fake neighbor is necessary for the modification of the classification result in the targeted attack mode. }
  \resizebox{0.45\textwidth}{!}{
          \begin{tabular}{c|c|c|c}
                \toprule \toprule
            Dataset           
            & Budget=1                       
            & Budget=2                     
            & Bud.1 + Bud.2 /\# Selected
            \\
            \midrule
                Cora
                    & 229
                    & 47
                    & 276/276
                    \\
                Citeseer 
                    & 199
                    & 46
                    & 245/245
                    \\
                CoraML 
                    & 269
                    & 44
                    & 313/314
                    \\
                Pubmed 
                    & 511
                    & 107
                    & 618/692
                    \\
                \bottomrule
                \bottomrule

          \end{tabular}
  }
\label{hyper2}%
\vspace{-3ex}
\end{table}%

\subsubsection{Surrogate model}
% We also conducted a set of experiments to compare proxy models, particularly focusing on the use of SGC as a proxy model. 
We use GCN as the surrogate model in this work. Several works have utilized SGC as a representative surrogate model~\cite{SGA}. The comparison results between the two surrogate models are shown in Table \ref{hyper: surrogate}. 
It is evident that in the majority of cases, using GCN as a proxy model yields significantly better attack effects compared to SGC as a proxy model. However, when the victim model itself is SGC, there are occasional instances where using SGC as a proxy model outperforms GCN as a proxy model. This occurrence is not surprising since it implies that the attacker possesses prior knowledge of the victim's model structure.  

\begin{table}[t]
  \caption{Comparison of attack effects between using different surrogate models (GCN and SGC)}
  \resizebox{0.47\textwidth}{!}{
          \begin{tabular}{c|c|ccccc}
                \toprule \toprule
        Dataset 
        & Surrogate                  
        & GCN                       
        & GAT                     
        & SGC                         
        & GraphSAGE                         
        & APPNP
        \\
        \midrule
        \multirow{2}{*}{Cora} 
        & SGC 
            & 72.52±0.46 % GCN
            & 73.52±0.65 % GAT
            & 66.64±0.33 % SGC
            & 73.14±0.36 % SAGE
            & 76.86±0.54 % APPNP
            \\
        & GCN 
            & 66.26±0.59 % GCN
            & 71.22±0.42 % GAT
            & 67.50±0.21 % SGC
            & 71.42±0.67 % SAGE
            & 74.60±0.64 % APPNP
            \\
        \midrule
        \multirow{2}{*}{Citeseer} 
        & SGC 
            & 65.26±0.45 % GCN
            & 65.30±0.51 % GAT
            & 63.98±0.25 % SGC
            & 66.36±0.45 % SAGE
            & 66.00±0.38 % APPNP
            \\
        & GCN 
            & 61.12±0.97 % GCN
            & 62.86±0.65 % GAT
            & 62.46±1.15 % SGC
            & 64.12±0.60 % SAGE
            & 64.14±0.96 % APPNP
            \\   
        \midrule
        \multirow{2}{*}{CoraML} 
        & SGC 
            & 72.38±0.83 % GCN
            & 74.90±0.40 % GAT
            & 61.80±0.11 % SGC
            & 73.22±0.49 % SAGE
            & 78.96±0.90 % APPNP
            \\
        & GCN 
            & 62.62±0.29 % GCN
            & 70.16±1.08 % GAT
            & 64.24±0.62 % SGC
            & 69.42±0.66 % SAGE
            & 71.96±0.71 % APPNP
            \\
        \midrule
        \multirow{2}{*}{Pubmed} 
        & SGC 
            & 39.92±0.66 % GCN
            & 42.64±2.84 % GAT
            & 31.70±0.71 % SGC
            & 54.42±1.01 % SAGE
            & 55.06±2.12 % APPNP
            \\
        & GCN 
            & 29.98±0.40 % GCN
            & 37.36±1.40 % GAT
            & 35.58±1.38 % SGC
            & 47.68±2.21 % SAGE
            & 43.12±2.84 % APPNP
            \\
                
            \midrule
            \midrule

    Dataset 
        & Surrogate                  
        & RGCN                       
        & MedianGCN                     
        & Jaccard                         
        & Grand                         
        & GNNGuard
        \\
        \midrule
        \multirow{2}{*}{Cora} 
        & SGC 
            & 72.92±0.55 % RGCN
            & 78.02±0.64 % Median
            & 75.82±1.06 % Jaccard
            & 79.14±0.56 % Grand
            & 77.96±0.10 % GNNGuard
            \\
        & GCN 
            & 68.08±1.00 % RGCN
            & 74.88±0.57 % Median
            & 72.80±0.98 % Jaccard
            & 76.58±0.33 % Grand
            & 76.08±0.21 % GNNGuard
            \\
        \midrule
        \multirow{2}{*}{Citeseer} 
        & SGC 
            & 65.40±0.40 % RGCN
            & 67.54±0.94 % Median
            & 68.56±0.46 % Jaccard
            & 70.00±0.40 % Grand
            & 68.48±0.49 % GNNGuard
            \\
        & GCN 
            & 61.14±0.75 % RGCN
            & 65.82±0.62 % Median
            & 67.62±0.46 % Jaccard
            & 68.98±0.68 % Grand
            & 67.66±0.53 % GNNGuard
            \\   
        \midrule
        \multirow{2}{*}{CoraML} 
        & SGC 
            & 71.72±0.46 % RGCN
            & 79.70±0.38 % Median
            & 74.90±0.63 % Jaccard
            & 79.46±0.53 % Grand
            & 77.10±0.60 % GNNGuard
            \\
        & GCN 
            & 63.74±0.12 % RGCN
            & 75.86±1.08 % Median
            & 71.94±0.55 % Jaccard
            & 75.96±0.89 % Grand
            & 76.34±0.49 % GNNGuard
            \\
        \midrule
        \multirow{2}{*}{Pubmed} 
        & SGC 
            & 39.12±1.59 % RGCN
            & 51.44±1.36 % Median
            & 51.16±0.47 % Jaccard
            & 53.86±1.23 % Grand
            & 62.60±1.21 % GNNGuard
            \\
        & GCN 
            & 31.84±0.71 % RGCN
            & 46.26±1.16 % Median
            & 41.56±1.76 % Jaccard
            & 45.60±1.97 % Grand
            & 56.08±3.21 % GNNGuard
            \\
                
            \bottomrule
            \bottomrule

          \end{tabular}
  }
\label{hyper: surrogate}%
\vspace{-3ex}
\end{table}%

\section{Conclusion}
In this work, we proposed a partial graph attack method called \mymodel{} for a more efficient graph global attack. As far as we know, \mymodel{} is the first method that takes into account the robustness of the victim nodes and directly selects vulnerable items as the attack targets. Rather than attacking all nodes, the partial attack can be more efficient under fixed attack budgets. With a hierarchical target selection policy and cost-efficient anchor-picking policy, \mymodel{} is capable of selecting vulnerable nodes and picking out the most promising anchors. Then, we proposed an iterative greedy-based attack method to perform efficient attacks. To validate the performance of \mymodel{}, we conducted comprehensive experiments on four datasets and targeted ten widely adopted graph neural network models, including both normal and robust GNN models. Experimental results demonstrate that \mymodel{} can achieve significant improvements in both attack effect and attack efficiency compared to other existing graph global attack methods. We believe that such a graph adversarial attack method could make further progress in the future from a new perspective.

\section*{Acknowledgment}
This work was supported by the National Natural Science Foundation of China (\#62102177), the Natural Science Foundation of Jiangsu Province (\#BK20210181), the Key R\&D Program of Jiangsu Province (\#BE2021729), Open Research Projects of Zhejiang Lab (\#2022PG0AB07), and the Collaborative Innovation Center of Novel Software Technology and Industrialization, Jiangsu, China.
%

% {\appendix[Proof of the Zonklar Equations]
% Use $\backslash${\tt{appendix}} if you have a single appendix:
% Do not use $\backslash${\tt{section}} anymore after $\backslash${\tt{appendix}}, only $\backslash${\tt{section*}}.
% If you have multiple appendixes use $\backslash${\tt{appendices}} then use $\backslash${\tt{section}} to start each appendix.
% You must declare a $\backslash${\tt{section}} before using any $\backslash${\tt{subsection}} or using $\backslash${\tt{label}} ($\backslash${\tt{appendices}} by itself
%  starts a section numbered zero.)}


% \section{References Section}
% You can use a bibliography generated by BibTeX as a .bbl file.
%  BibTeX documentation can be easily obtained at:
%  http://mirror.ctan.org/biblio/bibtex/contrib/doc/
%  The IEEEtran BibTeX style support page is:
%  http://www.michaelshell.org/tex/ieeetran/bibtex/

\bibliographystyle{IEEEtran}
\bibliography{PGA_TKDE}
 
\begin{IEEEbiography}[{\includegraphics[width=1in,height=1.25in,clip,keepaspectratio]{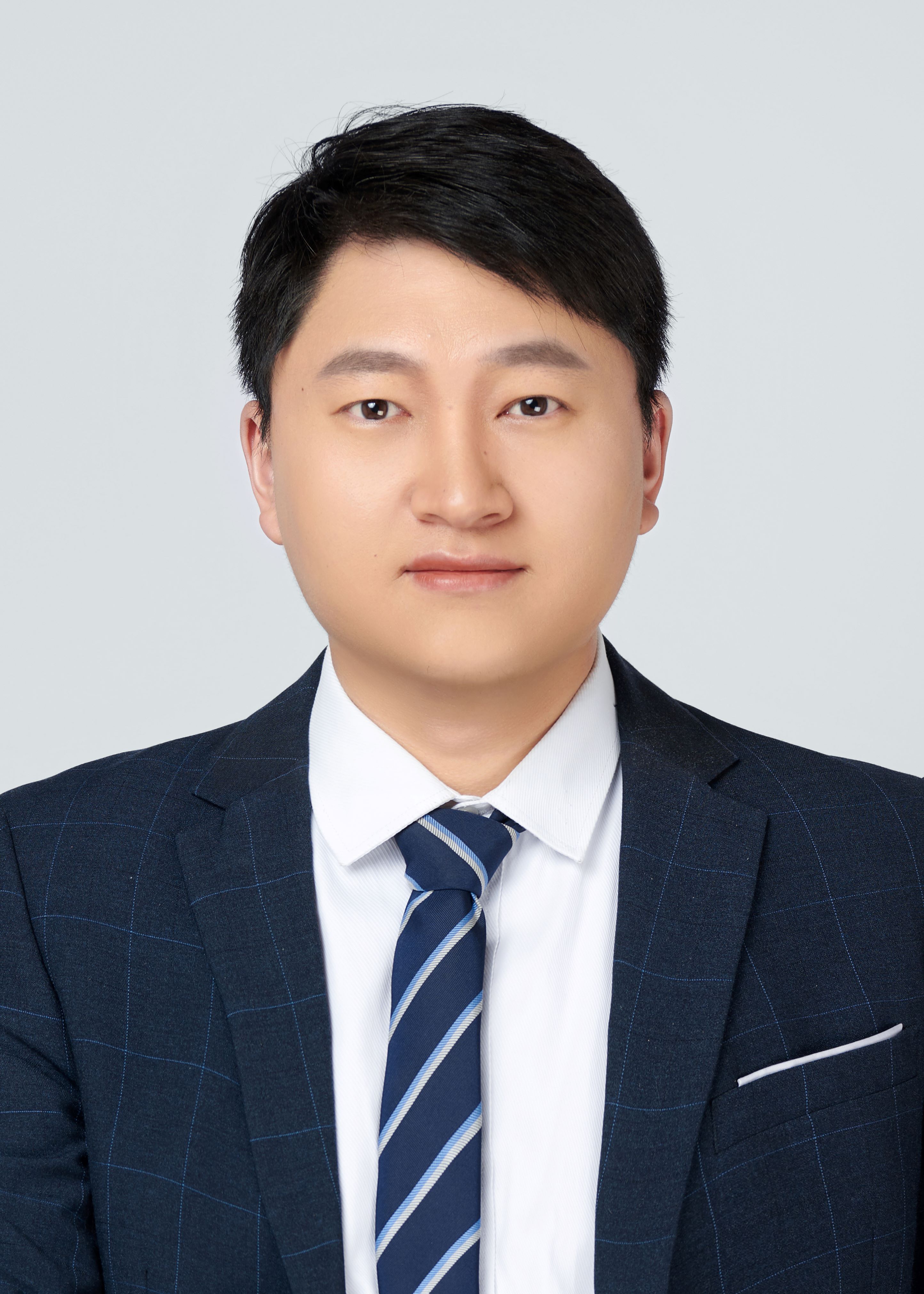}}]{Guanghui Zhu}
s currently an assistant researcher in the Department of Computer Science and Technology, and State Key Laboratory for Novel Software Technology, Nanjing University, China. He received his Ph.D. degrees in computer science and technology from Nanjing University. His main research interests include big data intelligent analysis, graph machine learning, and automated machine learning.
\end{IEEEbiography}
\begin{IEEEbiography}[{\includegraphics[width=1in,height=1.25in,clip,keepaspectratio]{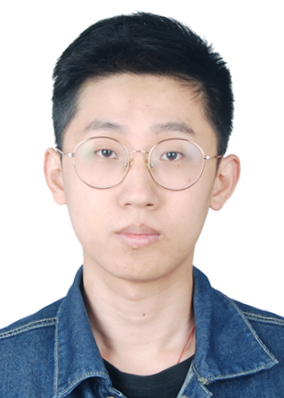}}]{Mengyu Chen}
is a postgraduate student in the Department of Computer Science and Technology, Nanjing University, China. He received his BS degree from Jiangnan University, China. His research interests include machine learning, graph neutral network, and graph adversarial
attack.
\end{IEEEbiography}
\begin{IEEEbiography}[{\includegraphics[width=1in,height=1.25in,clip,keepaspectratio]{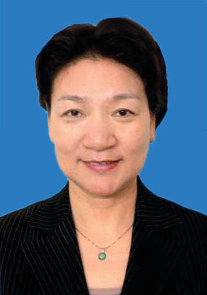}}]{Chunfeng Yuan}
is currently a professor in the Department of Computer Science and Technology, and State Key Laboratory for Novel Software Technology, Nanjing University, China. She received her bachelor and master degrees in computer science and technology from Nanjing University. Her main research interests include computer architecture, parallel and distributed computing, and information retrieval.
\end{IEEEbiography}
\begin{IEEEbiography}[{\includegraphics[width=1in,height=1.25in,clip,keepaspectratio]{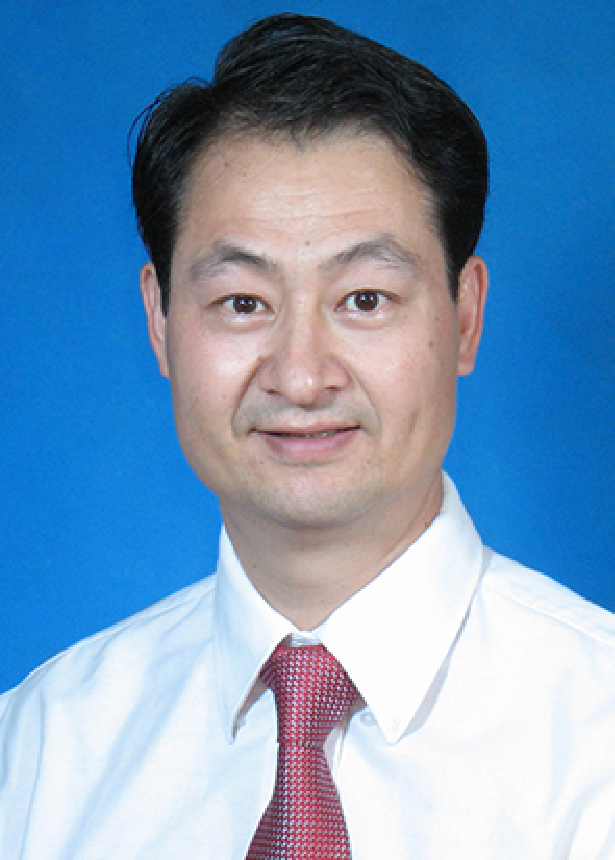}}]{Yihua Huang}
is currently a professor in the Department of Computer Science and Technology, and State Key Laboratory for Novel Software Technology, Nanjing University, China. He received his bachelor, master and Ph.D. degrees in computer science and technology from Nanjing University. His research interests include parallel and distributed computing, big data parallel processing, big data machine learning algorithm and system.
\end{IEEEbiography}
\end{document}